\documentclass[accepted]{uai2025} 
                        
\usepackage[american]{babel}

\usepackage{natbib} 
\usepackage{mathtools} 
\usepackage{booktabs} 
\usepackage{tikz} 

\usepackage[toc,page,header]{appendix}
\usepackage{minitoc}

\usepackage{times}
\usepackage{latexsym}
\usepackage{multirow}
\usepackage{array}

\usepackage[T1]{fontenc}

\usepackage[utf8]{inputenc}
\usepackage{booktabs}

\usepackage{microtype}

\usepackage{inconsolata}

\usepackage{subcaption}
\usepackage{graphicx}
\usepackage{enumitem}
\usepackage{amsmath}

\usepackage{algorithm}
\usepackage{algpseudocode}
\usepackage{xcolor}

\DeclareMathOperator*{\argmax}{arg\,max}

\title{The Consistency Hypothesis in Uncertainty Quantification \\ for Large Language Models}

\author[1]{Quan Xiao\textsuperscript{*}}
\author[2]{Debarun Bhattacharjya}
\author[2]{Balaji Ganesan}
\author[2]{Radu Marinescu}
\author[3]{Katsiaryna Mirylenka}
\author[2]{Nhan~H~Pham}
\author[2]{Michael Glass}
\author[2]{Junkyu Lee}

\makeatletter
  \renewcommand\AB@affilsepx{\quad}
\makeatother

\affil[1]{%
    Rensselaer Polytechnic Institute
}
\affil[2]{%
    IBM Research
}
\affil[3]{%
    Zalando
}
  
\begin{document}
\maketitle

\footnotetext{\textsuperscript{*} This work was conducted when QX was an intern and KM was a researcher at IBM Research. Corresponding authors: QX (\href{mailto:quanx1808@gmail.com}{quanx1808@gmail.com}), DB (\href{mailto:debarunb@us.ibm.com}{debarunb@us.ibm.com}).} 

\vspace{-0.3cm}

\begin{abstract}
Estimating the confidence of large language model (LLM) outputs is essential for real-world applications requiring high user trust. Black-box uncertainty quantification (UQ) methods, relying solely on model API access, have gained popularity due to their practical benefits. In this paper, we examine the implicit assumption behind several UQ methods, which use generation consistency as a proxy for confidence—an idea we formalize as the \emph{consistency hypothesis}. We introduce three mathematical statements with corresponding statistical tests to capture variations of this hypothesis and metrics to evaluate LLM output conformity across tasks. Our empirical investigation, spanning 8 benchmark datasets and 3 tasks (question answering, text summarization, and text-to-SQL), highlights the prevalence of the hypothesis under different settings.
Among the statements, we highlight the `Sim-Any' hypothesis as the most actionable, and demonstrate how it can be leveraged by proposing data-free black-box UQ methods that aggregate similarities between generations for confidence estimation. These approaches can outperform the closest baselines, showcasing the practical value of the 
empirically observed consistency hypothesis. 
\end{abstract}

\section{Introduction \& Related Work}

Large language models (LLMs) have become pervasive due to their state-of-the-art performance for various natural language processing tasks. Despite recent advances, LLMs are known to suffer from important limitations that heavily impact their usability in diverse real-world applications, including lack of traceability, context sensitivity, and difficulty around incorporating 
domain-specific knowledge and handling rare queries as well as dynamic and evolving data. 
These issues often lead to unpredictable or unverifiable responses, or outputs that appear authoritative even when they are incorrect, which can undermine user trust in high-stakes domains such as healthcare, finance, and law.



\emph{Uncertainty quantification} (UQ) approaches provide insights into the reliability of an LLM's predictions by associating them with \emph{confidence} estimates, making them a critical component of many real-world systems.
The estimated confidences should ideally be well \emph{calibrated}, as gauged by the degree to which they match the empirical accuracy for that prediction~\citep{murphy1967verification,dawid1982well}.
They can be used to distinguish correct LLM responses from incorrect ones, as well as for \emph{selective generation}, i.e. for rejecting a fraction of the instances that one is least confident about~\citep{el2010foundations}. 
Robust UQ enables downstream decision-making systems to act conservatively when necessary, and can facilitate human-AI collaboration by allowing users to identify when to trust or question model outputs. Moreover, well-calibrated uncertainty estimates are essential for integrating LLMs into broader pipelines where resource allocation or fallback mechanisms depend on confidence-aware behavior.

There is a growing body of literature on UQ techniques for estimating the confidence of LLM generations.
\emph{Verbalized} UQ approaches rely on prompting LLMs to express uncertainty about a generated response in natural language, such as through qualifying phrases or numbers~\citep{lin2022teaching,kadavath2022language,mielke-etal-2022-reducing}.
For instance, \citet{kadavath2022language} use a self-verification approach for question answering tasks where an LLM is asked to declare whether a provided answer is true or false for a question, and the model's token logit is used to infer the probability that the answer is correct.
Although some empirical studies show that verbalized confidences using LLMs trained via the reinforcement learning with human feedback paradigm can
yield well-calibrated estimates~\citep{tian2023just}, others indicate that LLMs are overconfident when verbalizing their own confidence~\citep{xiong2024llms} and that LLMs' meta-linguistic judgments are less reliable than quantities derived directly from token-level probabilities~\citep{hu2023prompting}.


\begin{figure*}[!t]
\centering
\begin{subfigure}{0.24\textwidth}
\includegraphics[width=\textwidth]{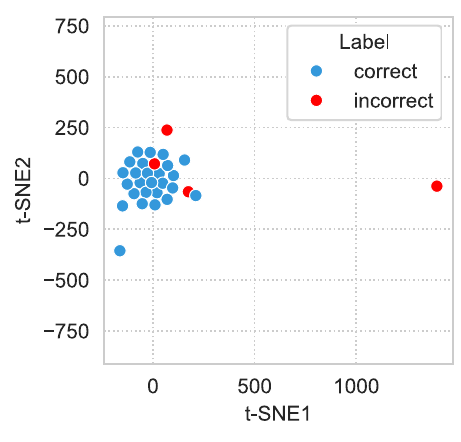}
\caption{QA case 1}
\label{fig:text1}
\end{subfigure}
\hfill
\begin{subfigure}{0.24\textwidth}
\includegraphics[width=\textwidth]{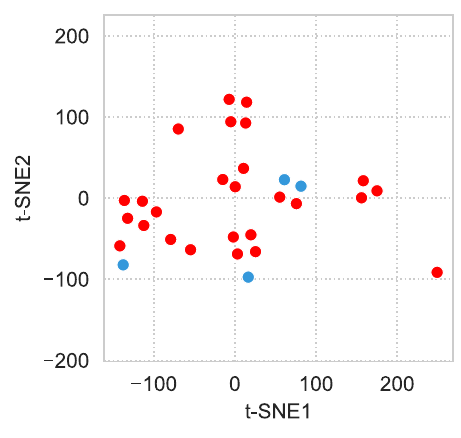}
\caption{QA case 2}
\label{fig:text2}
\end{subfigure}
\hfill
\begin{subfigure}{0.24\textwidth}
\includegraphics[width=\textwidth]{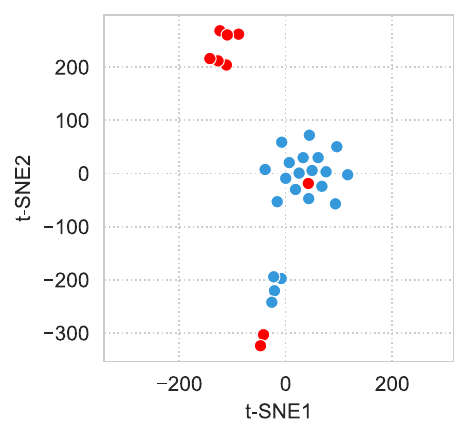}
\caption{Text-to-SQL case 1}
\label{fig:sql1}
\end{subfigure}
\hfill
\begin{subfigure}{0.24\textwidth}
\includegraphics[width=\textwidth]{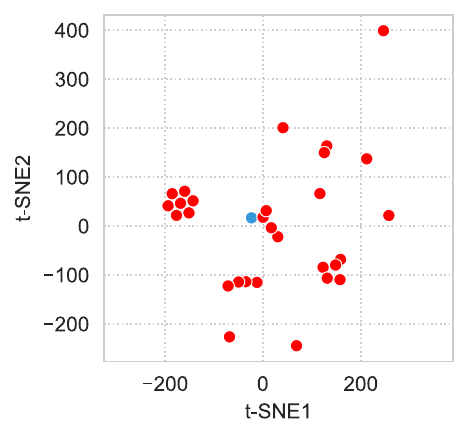}
\caption{Text-to-SQL case 2}
\label{fig:sql2}
\end{subfigure}
\caption{T-SNE projections of 30  generations represented by sentence embeddings, for 2 instances each from the CoQA and Spider Realistic dev datasets. Correct and incorrect generations are labeled in blue and red respectively. The plots suggest that correct generations tend to cluster closely to other generations and to each other as compared to incorrect generations, whereas incorrect generations are often more dispersed and exhibit greater dissimilarity. }
\label{fig:embeddings}
\vspace{-0.5cm}
\end{figure*}

UQ approaches, regardless of whether they are verbalized or not, can be categorized into two high-level categories -- 
\emph{black-box} techniques only assume API access to an LLM~\citep{lin2023generating,manakul2023selfcheckgpt,cole2023selectively}, whereas \emph{white-box} approaches require other model information such as the model weights 
\citep{gal16,xiao2019quantifying}, or internal states such as embeddings and activation spaces~\citep{ren2023outofdistribution},
or 
only token-level  logits~\citep{kuhn2022semantic,kadavath2022language}. 
Some recent white-box methods for calibrating LLM output typically require substantial training data for fine-tuning  LLMs~\citep{chen2023adaptation,kapoor-etal-2024-calibration,ulmer-etal-2024-calibrating}, making them computationally intensive. In comparison, black-box approaches offer practical advantages, including robustness to LLM upgrades, compatibility with proprietary models (even those limited to API-based inputs and outputs), and computational tractability at inference time. As a result, such approaches have gained popularity in generation tasks.

Several UQ approaches, both black-box and white-box, sample multiple generations from an LLM and characterize some notion of variability between generations as a proxy for confidence. 
For instance, \citet{kuhn2022semantic} find semantic equivalence between generations to cluster them and then compute the entropy of each cluster, where a lower entropy value implies a more confident generation.
\citet{lin2023generating} take a spectral clustering approach instead by treating generations as nodes in a graph, but also rely on variability. 
Various other approaches leverage consistency between generations via different forms of perturbation~\citep{spuq}, for addressing challenges such as ambiguous questions~\citep{cole2023selectively} or detecting hallucinated facts~\citep{manakul2023selfcheckgpt}.


Recent efforts have recommended combining ideas from the various UQ categories, such as an approach that estimates a numeric confidence score for any LLM output by combining an extrinsic consistency-based metric over multiple samples with an intrinsic confidence estimate obtained by prompting the LLM itself~\citep{chen2023quantifying}. 
\citet{xiong2024llms} suggest that using both verbalized confidence
and sampling consistency could be promising for attaining more accurate confidence assessments in LLMs.



There is an implicit assumption behind all approaches that generate multiple samples and use consistency or variability to infer confidence -- informally,
when a generated response is more different than others, it is more likely to be incorrect, thus responses that are consistently similar are more likely to be correct. We refer to this assumption as the \emph{consistency hypothesis}, and
visualize it using Figure~\ref{fig:embeddings}, which considers two instances each from two datasets -- CoQA~\citep{reddy2019coqa} for open-ended question answering and Spider Realistic~\citep{deng2021structure} for the text-to-SQL generation task.
We represent 30 generations for each instance via t-SNE projections of their semantic encodings, distinguishing between generations deemed to be correct and incorrect through color coding. For instances across both types of datasets, we observe that correct generations tend to be visually closer to other generations and to each other as compared to incorrect generations (Figures~\ref{fig:text1},~\ref{fig:sql1}). Incorrect generations tend to lie on the border of the correct generations. Furthermore, when there are many incorrect generations, these are often largely spread across the representation space and may be dissimilar (Figures~\ref{fig:text2},~\ref{fig:sql2}).
This is merely an illustrative visual examination for specific instances, but how can this crucial assumption that forms the basis for so many UQ approaches be formalized and tested? Importantly, \textbf{how can one measure the extent to which the consistency hypothesis holds for LLM generations for a dataset}?


In this paper, we continue the burgeoning investigation of consistency-based methods for UQ in LLMs. Our specific \textbf{contributions} are as follows:

\begin{itemize}[noitemsep,nolistsep,leftmargin=*]
\item We propose a rigorous statistical procedure (as outlined in Algorithm \ref{alg: fraction_group_num}) for testing the implicit assumption behind consistency-based UQ approaches for LLMs using  notions of 
similarity between generations. We propose both conceptual and actionable versions of the hypothesis stated verbally, and formulate them as comparisons between sets of similarities. 
\item We propose metrics to measure the extent to which various versions of the consistency hypothesis hold for LLM generations for a task and dataset. 
\item  We conduct a detailed empirical investigation spanning 8 benchmark datasets over 3 tasks – question answering, text summarization, and text-to-SQL
– 
to verify the extent to which the hypotheses hold for these datasets.
\item Based on the empirical findings, we design a black-box UQ method associated with new aggregation functions that performs reasonably well for confidence estimation, further demonstrating the practical value of exploiting the consistency hypothesis for UQ and motivating future exploration into related approaches. 
\item In addition, our illustrative visualization tools such as those depicted in Figures \ref{fig:embeddings} and \ref{fig:boxplot} can aid data exploration pertaining to the consistency hypothesis (see Appendix \ref{sec:dis_visualization} for further examples). 

\end{itemize}

\section{Consistency Hypothesis Formalization}

We begin by clarifying the problem setup with notation, and subsequently propose and formalize various versions of the consistency hypothesis.

\subsection{Preliminaries \& Notation}

Consider a dataset with queries $x_1, \cdots, x_n$, where $x_i$ is the $i^{th}$ query (a.k.a. instance, such as a question in a question answering task). 
Consistency-based approaches call an LLM for multiple generations (a.k.a. samples, such as answers in a question answering task) for each instance. Generations corresponding to $x_i$ are denoted
$y_i^1, \cdots, y_i^m$.
We use $[n]:= \{1, \cdots, n\}$ and $[m]:= \{1, \cdots, m\}$ as shorthand to denote the set of indices for queries and generations, respectively.

We assume there is a set of responses $\mathcal{Y}^*_i$ that are deemed \emph{correct} for query $x_i$. Therefore, a generated output $y_i^j$ is correct if $y_i^j \in \mathcal{Y}^*_i$.
For tasks such as open-ended question answering and summarization, a generation for a particular input may be viewed as correct if a metric such as Rouge-L with respect to the ground truth output exceeds some predetermined threshold.

We also assume access to a \emph{similarity} metric $s(\cdot)$ defined on the interval $[0,1]$, to compute pairwise similarities
$s_i^{j,k} = s(y_i^j, y_i^k)$, which is the similarity between the $j^{th}$ and $k^{th}$ samples for the $i^{th}$ instance in the dataset.  
In addition, an \emph{aggregation} function $f(\cdot)$ is required for combining pairwise similarities into a scalar. 




\subsection{Statements \& Formalizations}

\begin{figure}
\includegraphics[width=0.5\textwidth]{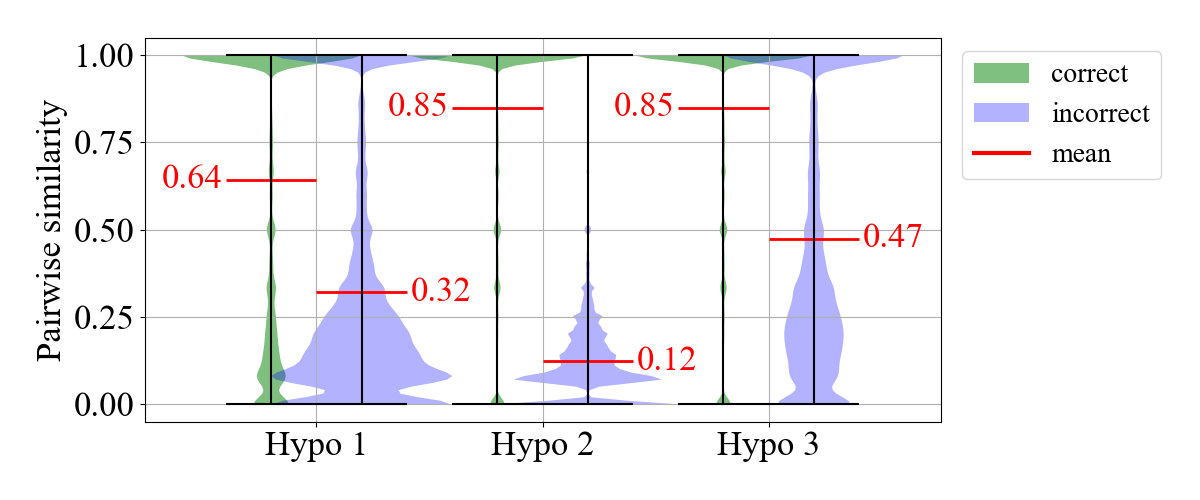}
\vspace{-0.8cm}
\caption{Visualization of the distributions of the Jaccard pairwise similarities of correct cluster $S^C$ and incorrect cluster $S^I$ under different hypotheses from Granite generations of $30$ samples each on the CoQA dataset. Complete visualizations for each dataset can be found in Figure \ref{fig:boxplot-2}. }
\label{fig:boxplot}
\vspace{-0.5cm}
\end{figure}

We propose three different verbal statements as candidates for the consistency hypothesis, and formalize them as mathematical statements. 
These formal statements could either be at the ``instance-level'', i.e. at the level of a particular query involving multiple generations, or at the ``group-level'', i.e.  applicable over a set of queries in the dataset. As a preview, we visualize the distributions of two similarity sets, denoted $S^C$ (correct cluster) and $S^I$ (incorrect cluster) in Figure \ref{fig:boxplot}; these are keys for formalizing the consistency hypothesis and will be defined subsequently. We highlight that these distributions are clearly distinct across all three verbal hypotheses proposed. Furthermore, the mean of the correct set notably exceeds that of the incorrect set, which will form the basis for our formal statements. 

Motivated by Figure \ref{fig:embeddings}, we first introduce two natural hypotheses that compare generations with either other correct generations or those within the same category. We then propose a more actionable hypothesis designed for the practical scenario where generation labels are unavailable. 

\subsubsection{
The Sim-Correct Hypothesis}

One of the natural verbal statements inspired by Figure \ref{fig:embeddings} is as follows: \textit{A correct generation is more similar to other correct generations for an instance, as compared to an incorrect generation}. We call this statement the \emph{sim-correct} hypothesis, as it makes a claim about similarity only with respect to \emph{correct} generations. 

We consider two variants for formalization:  one considers similarity of a generation with respect to other individual generations, and the other aggregates similarities with respect to other generations. 

\noindent\textbf{Pairwise Consistency.}
First we consider pairwise similarities between generations. 
We build two sets of pairwise similarities for each instance; one is for correct generations and one for incorrect generations: $S_i^C = \{s_i^{j,k}: j,k \in \mathcal{Y}^*_i, k \neq j \}$ and
$S_i^I = \{s_i^{j,k}: j \notin \mathcal{Y}^*_i, k \in \mathcal{Y}^*_i  \}$ where $k \in [m]$. We construct these sets by selecting the appropriate entries from the matrix formed by computing pairwise similarities between all generations. 

The formal statement can now be made as: $\mu_i^C > \mu_i^I$, where these are the means of sets $S_i^C$ and $S_i^I$ respectively.
The statement could be verified using a one-sided t-test by rejecting the null hypothesis of $\mu_i^C \leq \mu_i^I$ at a suitable $p$-value. 

Note that the above statements are instance-level as there is a statement for each $i$-th instance. There may however not be enough data for statistical significance at this level.
Group level versions of the sets can be obtained by collecting over  all instances in a group: $S_g^C = \bigcup_{i \in G} \{s_i^{j,k}: j,k \in \mathcal{Y}^*_i, k \neq j \}$ and
$S_g^I = \bigcup_{i \in G} \{s_i^{j,k}: j \notin \mathcal{Y}^*_i, k \in \mathcal{Y}^*_i  \}$, where group $g$ is associated with indices of queries $G \subseteq [n]$ and $k \in [m]$.
The formal statement at the group level is $\mu_g^C > \mu_g^I$. When $G = [n]$, there is only 1 group and the statement is at the ``dataset level''. We denote the correct and incorrect sets at the dataset level as $S^C$ and $S^I$, respectively.


\noindent\textbf{Aggregated Consistency.}
Here we consider the aggregated similarity over all other samples. Besides  similarity metric $s(\cdot)$, this requires an aggregation function $f(\cdot)$. 
In this case, we use aggregated similarities instead of pairwise ones to gauge how a generation compares with other generations:
$S_i^C = \{\bar{s}_i^j: j \in \mathcal{Y}^*_i \}$ and
$S_i^I = \{\bar{s}_i^j: j \notin \mathcal{Y}^*_i\}$, where $\bar{s}_i^j = f(s_i^{j,i_1}, \cdots, s_i^{j,i_m})$ with $\{s_i^{j,i_1}, \cdots, s_i^{j,i_m}\}\in\mathcal{Y}_i^*$ is the aggregated similarity of generation $j$ w.r.t other correct generations for instance $i$. 
The formal statement remains $\mu_i^C > \mu_i^I$ using means for the aforementioned sets, and
once again 
one can consider either instance-level statements or those collecting over instances in a group.

The pairwise consistency hypothesis offers more data points for conducting statistical tests, while the aggregated consistency hypothesis is more practical for designing new UQ methods with improved aggregation functions.

\subsubsection{
The Sim-Separate Hypothesis}

According to Figure \ref{fig:embeddings}, another candidate verbal statement is: \textit{A correct generation is more similar to  other correct generations than an incorrect generation is to other incorrect generations, for an instance}.
We refer to this statement as the \emph{sim-separate} hypothesis as its assertion entirely separates correct and incorrect generations. 

The main difference between the formalization for this hypothesis and the previous one is how instance-level similarity sets are defined. Here, 
$S_i^C = \{s_i^{j,k}: j,k \in \mathcal{Y}^*_i, k \neq j \}$ and
$S_i^I = \{s_i^{j,k}: j,k \notin \mathcal{Y}^*_i, k \neq j \}$ where $k \in [m]$. Note that the aggregated consistency version of this hypothesis involves aggregating similarities for a correct generation over the set of all correct generations, and for an incorrect generation over the set of all incorrect generations, for a given instance. All other aspects are the same as Sim-Correct.

\subsubsection{
The Sim-Any Hypothesis}

When labels for correct and incorrect generations are unavailable, as is typical, one black-box UQ approach for confidence estimation \citep{lin2023generating} is to leverage this aggregation function directly, e.g. confidence in $y_i^j$ could be computed as 
$c_i^j = f(s_i^{j,1}, \cdots, s_i^{j,m})$ and 
interpreted as the probability of the generation being correct, potentially after further calibration. Thus, while conceptually intuitive, the previous two hypotheses are insufficient for providing practical guidance in designing black-box UQ methods. We then propose a novel formalization of the consistency hypothesis by comparing each generation with all other generations, eliminating the need for labels during aggregation. 

A verbal statement for this version of the consistency hypothesis is as follows:
\textit{A correct generation is more similar to other generations for an instance, as compared to an  incorrect generation}. We refer to this statement as the \emph{sim-any} hypothesis, as the statement makes a claim about similarity with respect to \emph{any} other generation or generations, not necessarily those that are correct or incorrect.

For this hypothesis, instance-level similarity sets are constructed as: 
$S_i^C = \{s_i^{j,k}: j \in \mathcal{Y}^*_i, k \neq j \}$ and
$S_i^I = \{s_i^{j,k}: j \notin \mathcal{Y}^*_i, k \neq j \}$ where $k \in [m]$.
The aggregated consistency version of this hypothesis performs aggregation only over the set of all generations for an instance;
All other aspects remain the same as Sim-Correct. As the Sim-Any hypothesis is less intuitive from Figure \ref{fig:embeddings}, it should be carefully examined by statistical tests. If verified to some extent, 
the confidence in a generation could potentially be estimated by aggregating its similarity with all other generations.

In subsequent sections (including figures), we also refer to the Sim-Any, Sim-Correct, and Sim-Separate hypotheses as Hypo \#1, \#2, and \#3, respectively, for brevity.

\section{Consistency Hypothesis Verification}

How can one measure the extent to which a dataset conforms to a consistency hypothesis? We propose the following metrics that leverage the formal statements. 
The metrics rely on Algorithm \ref{alg: fraction_group_num}, which is provided the 
number of groups $n_g$ and returns the fraction of groups that verify the hypothesis, denoted $\rho(n_g)$, as determined by whether the null hypothesis is rejected at $p$-value $\leq 0.05$.




\begin{algorithm}[t]
\caption{Fraction of verified groups  $\rho(n_g)$} 
\begin{algorithmic}[1]
\State \textbf{Input:} Similarity matrix of generations, number of groups $n_g$, number of repetitions $R$
\For{each repetition $r$ from 1 to $R$}
    \State Split dataset randomly into $n_g$ groups
    \State Initialize $n_v \gets 0$
    \For{each group $g$ in $n_g$}
        \State Construct similarity sets $S_g^C$ and $S_g^I$
        \State Conduct a t-test on the means 
        \State Calculate the $p$-value
        \If{$p$-value $\leq 0.05$}
            \State $n_v\gets n_v+1$
        \EndIf
    \EndFor
    \State Compute $\rho_r(n_g) \gets \frac{n_v}{n_g}$
\EndFor
\State Compute $\rho(n_g) \gets \frac{1}{R}\sum_{r=1}^R \rho_r (n_g)$
\State Compute the error bar of $\rho(n_g)$ as the variance 
\State \textbf{Output:} $\rho(n_g)$ with the error bar  
\end{algorithmic}
\label{alg: fraction_group_num}
\end{algorithm} 

\begin{figure*}[htb]
\setlength{\tabcolsep}{-0.05cm}
\begin{tabular}{ccc}
\includegraphics[width=.35\textwidth]{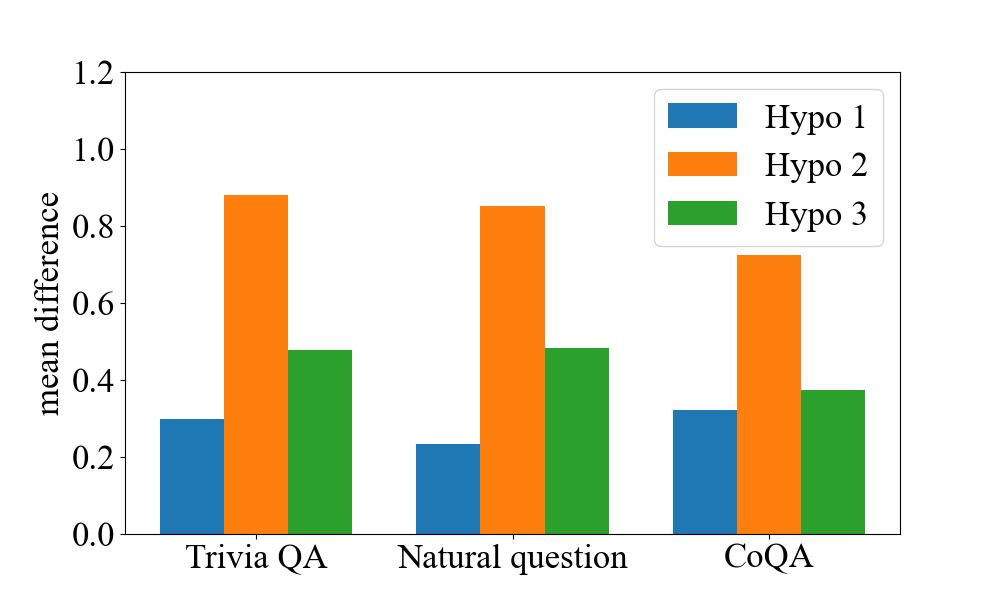}
&
\includegraphics[width=.35\textwidth]{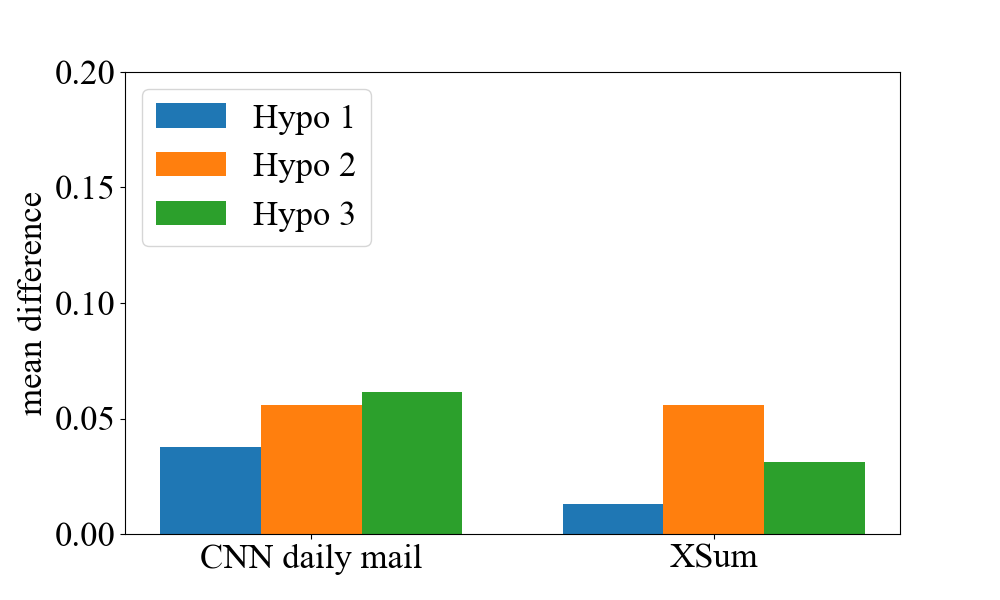}
&\includegraphics[width=.35\textwidth]{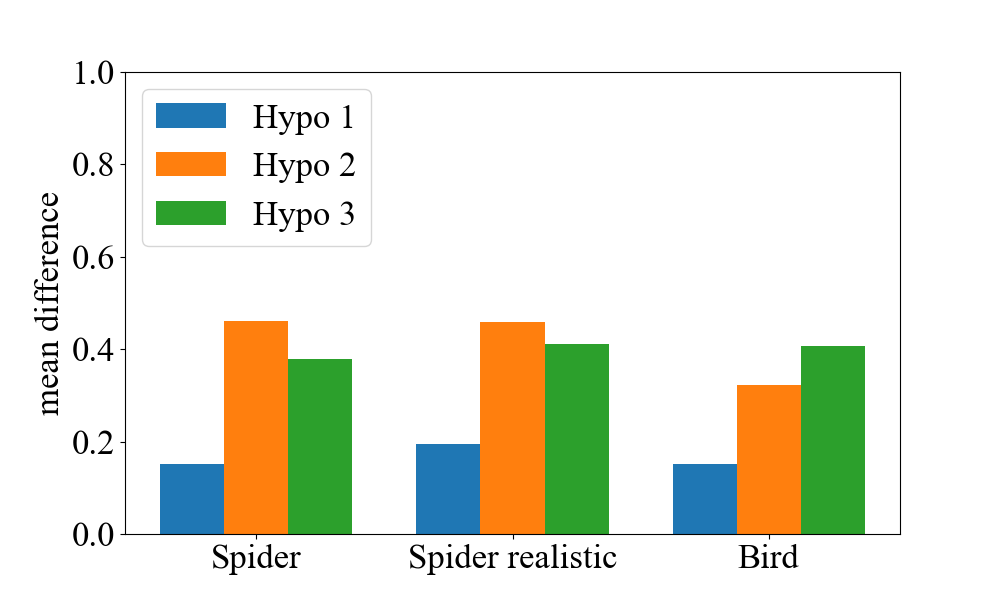}\\
{\footnotesize(a) QA} & {\footnotesize(b) Summarization }& {\footnotesize(c) Text-to-SQL}
\end{tabular}
\vspace{-0.2cm}
 \caption{Verification of hypotheses using mean difference $\Delta \mu$  between similarity sets with Jaccard pairwise similarity on all $8$ datasets for the QA, summarization, and text-to-SQL tasks. }
\label{fig:hypo_mean_diff} 
\vspace{-0.4cm}
\end{figure*}

\begin{figure*}[htb]
\setlength{\tabcolsep}{-0.05cm}
\begin{tabular}{ccc}
\includegraphics[width=.35\textwidth]{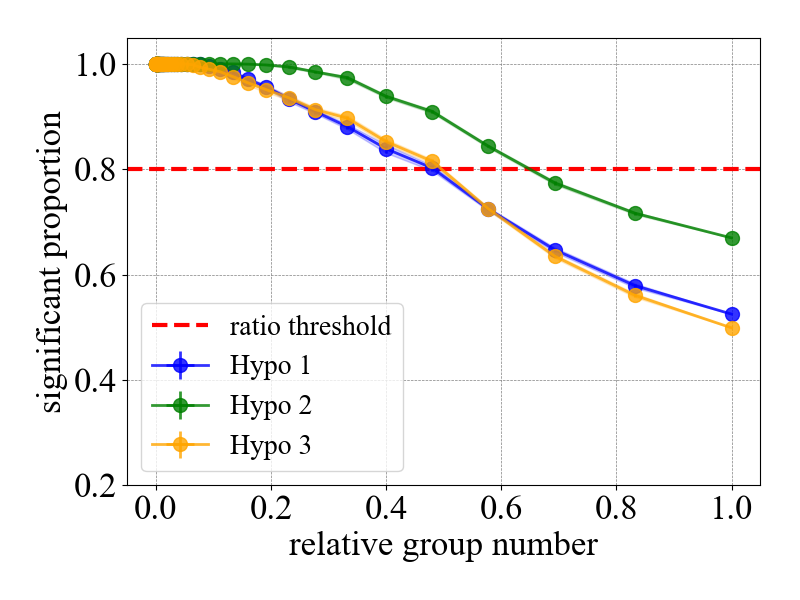}
&
\includegraphics[width=.35\textwidth]{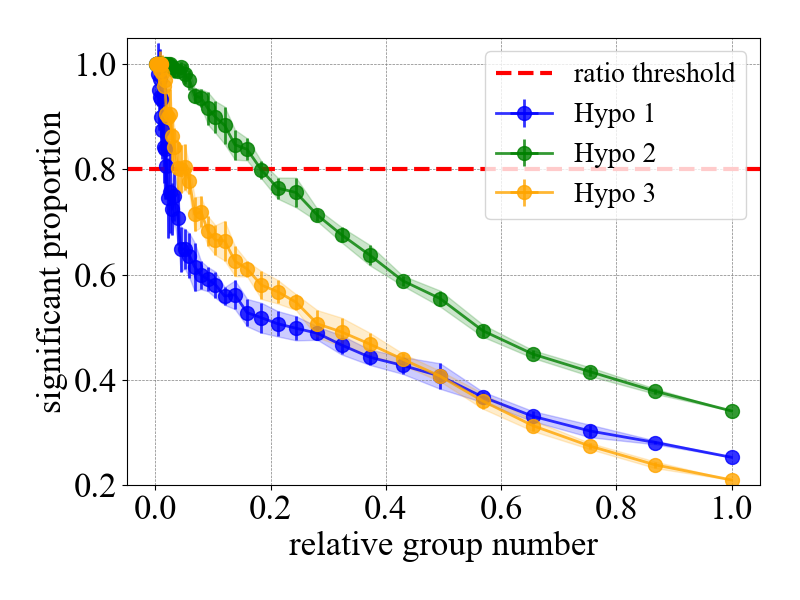}
&\includegraphics[width=.35\textwidth]{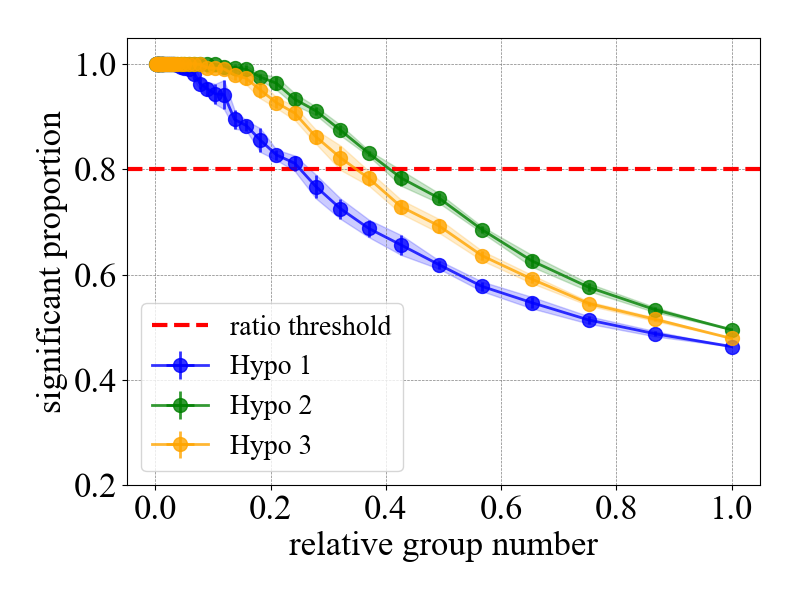}\\
{\footnotesize(a) CoQA dataset} & {\footnotesize(b) XSum dataset }& {\footnotesize(c) Spider dataset}
\end{tabular}
\vspace{-0.2cm}
 \caption{Verification of hypotheses (pairwise consistency statements) using the fraction of verified groups $\rho (n_g)$ as a function of the relative number of groups $n_g/n$ with Jaccard pairwise similarity on representative datasets for QA, summarization, and text-to-SQL tasks. Complete results for $8$ datasets are included in the Appendix. } 
\label{fig:hypo_jaccard_pairwise_main_paper} 
\vspace{-0.4cm}
\end{figure*}

\noindent\textbf{Mean difference} $\Delta \mu = \mu^C - \mu^I$. This measures the extent to which the statement holds true across the entire dataset but is problematic as a measure of the efficacy of consistency-based UQ methods, which necessarily need to be applied at the instance level; this is because it compiles similarity sets using all instances and only applies at the dataset level. There is likely to be sufficient data to test the hypothesis at this level. 

\noindent\textbf{Fraction of instances where the hypothesis is verified} $\rho(n)$. This is an instance level measure that calls upon Algorithm~\ref{alg: fraction_group_num} with a group for each instance, i.e. $n_g=n$.
Note that there may not be sufficient data at the instance level for  testing the statistical significance of a statement.  

\noindent\textbf{Maximum relative number of groups for a thresholded fraction of  verified groups}:
$\theta^* = n_g^*/n $ for
$n_g^* = \argmax_{n_g} {I (\rho (n_g) \geq \rho^* )}$, where $I(\cdot)$ is the indicator function, $\rho(n_g)$ is the fraction of verified groups obtained from Algorithm~\ref{alg: fraction_group_num}, $\rho^*$ is some user-specified threshold $\%$, and $n$ is the number of instances in the dataset. We use the relative number of groups $\theta = n_g/n$ for comparability across datasets. 
This metric is intended to balance the other two metrics by considering both statistical power (significant proportion) and practicality (relative group number) for the purpose of conducting consistency-based UQ. 
Moreover, this metric is a summary quantity that effectively represents the `height' of the trade-off curve of statistical power and practicality using a single measure — a higher curve means that a hypothesis is more valid for that dataset. Although we selected  $\rho^*=80\%$ as a  threshold in experiments for quantification purposes, our conclusions are based on a comparative analysis of the curves rather than hinging upon this specific value. 

These metrics serve as tools for evaluating the verification extent of different consistency hypotheses based on various similarity metrics $s(\cdot)$ and aggregation functions $f(\cdot)$. They provide essential building blocks for comparing metric performance and inspiring the development of new black-box UQ methods based on the verification extent.

\section{Empirical Investigation}
\label{sec:empirical}

We conduct experiments for various generative tasks, covering several datasets and generations with representative LLMs to tackle those datasets, with the intent of understanding the extent to which the hypotheses hold and how they could be leveraged for UQ. 

\begin{table*}
\centering
\resizebox{\linewidth}{!}{%
\begin{tabular}{lcccccccccl}\toprule
\multirow{2}{*}{\shortstack{Aggregation/\\Similarity}} & \multicolumn{4}{c}{Mean difference when $n_g=10$ $\uparrow$} & \multicolumn{4}{c}{Max relative group number when $\rho^*=0.6$ $\uparrow$ } \\
\cmidrule(lr){2-5} \cmidrule(lr){6-9} 
& Jaccard & Rouge-1 & Rouge-L & Sbert & Jaccard & Rouge-1 & Rouge-L & Sbert \\\midrule
\shortstack{pairwise \\ \vspace{4pt}}    
& \shortstack{$0.2247$ \\ ${\scriptstyle\pm 0.0009}$} & \shortstack{$0.2329$ \\ ${\scriptstyle\pm 0.0008}$} & \shortstack{$\bf 0.2331$ \\ ${\scriptstyle\pm \bf 0.0007}$} & \shortstack{$0.1713$ \\ ${\scriptstyle\pm 0.0005}$} & \shortstack{$0.1838$ \\ ${\scriptstyle\pm 0.00002}$} & \shortstack{$0.1992$ \\ ${\scriptstyle\pm 0.00002}$} & \shortstack{$0.1994$ \\ ${\scriptstyle\pm 0.00003}$} & \shortstack{$\bf 0.2194$ \\ ${\scriptstyle\pm \bf 0.00003}$} \\
\shortstack{arithmetic \\ \vspace{4pt}}  
& \shortstack{$0.0816$ \\ ${\scriptstyle\pm 0.0011}$} & \shortstack{$0.0897$ \\ ${\scriptstyle\pm 0.0010}$} & \shortstack{$0.0898$ \\ ${\scriptstyle\pm 0.0012}$} & \shortstack{$0.0858$ \\ ${\scriptstyle\pm 0.0005}$} & \shortstack{$0.0464$ \\ ${\scriptstyle\pm 0.00009}$} & \shortstack{$0.0587$ \\ ${\scriptstyle\pm 0.00003}$} & \shortstack{$0.0548$ \\ ${\scriptstyle\pm 0.00005}$} & \shortstack{$0.1096$ \\ ${\scriptstyle\pm 0.00012}$} \\
\shortstack{geometric \\ \vspace{4pt}} 
& \shortstack{$0.1437$ \\ ${\scriptstyle\pm 0.0010}$} & \shortstack{$0.1729$ \\ ${\scriptstyle\pm 0.0009}$} & \shortstack{$0.1729$ \\ ${\scriptstyle\pm 0.0011}$} & \shortstack{$0.1127$ \\ ${\scriptstyle\pm 0.0004}$} & \shortstack{$0.1176$ \\ ${\scriptstyle\pm 0.00002}$} & \shortstack{$0.1453$ \\ ${\scriptstyle\pm 0.00001}$} & \shortstack{$0.1450$ \\ ${\scriptstyle\pm 0.00001}$} & \shortstack{$0.1532$ \\ ${\scriptstyle\pm 0.00002}$} \\
\shortstack{harmonic \\ \vspace{4pt}}  
& \shortstack{$0.1417$ \\ ${\scriptstyle\pm 0.0010}$} & \shortstack{$0.1722$ \\ ${\scriptstyle\pm 0.0011}$} & \shortstack{$0.1722$ \\ ${\scriptstyle\pm 0.0008}$} & \shortstack{$0.1425$ \\ ${\scriptstyle\pm 0.0004}$} & \shortstack{$0.0410$ \\ ${\scriptstyle\pm 0.00002}$} & \shortstack{$0.1468$ \\ ${\scriptstyle\pm 0.00007}$} & \shortstack{$0.1536$ \\ ${\scriptstyle\pm 0.00003}$} & \shortstack{$0.1714$ \\ ${\scriptstyle\pm 0.00009}$} \\
\bottomrule
\end{tabular}
}
\caption{Ablation study on choice of similarity metric and aggregation functions for the Natural Question dataset. Higher numbers indicate stronger verification of the consistency hypothesis H1.}
\vspace{-0.3cm}
\label{table:sim_agg_study_natural_question}
\end{table*}

\begin{figure*}[htb]
\setlength{\tabcolsep}{-0.05cm}
\begin{tabular}{ccc}
\includegraphics[width=.35\textwidth]{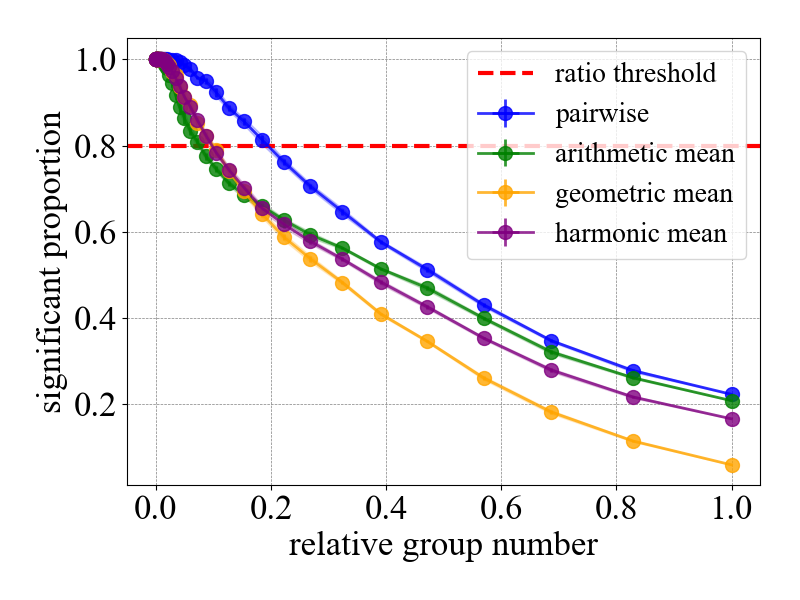}
&
\includegraphics[width=.35\textwidth]{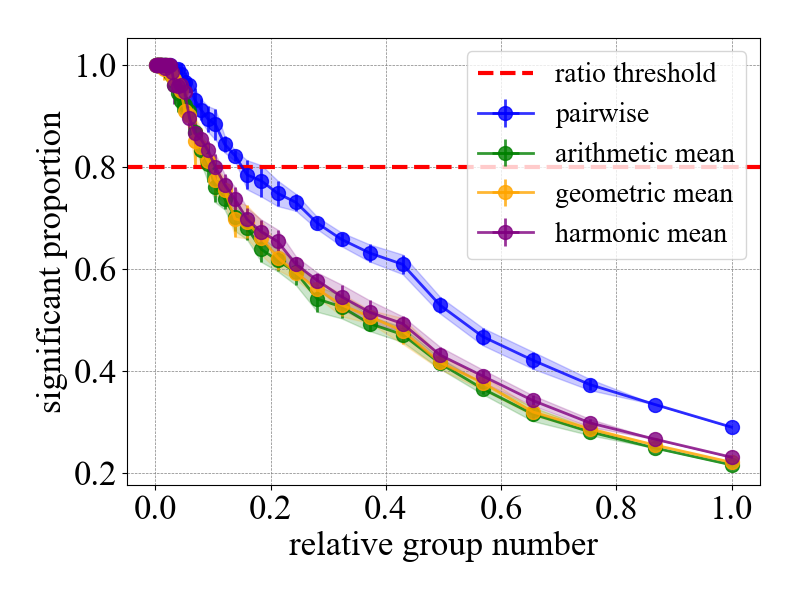}
&\includegraphics[width=.35\textwidth]{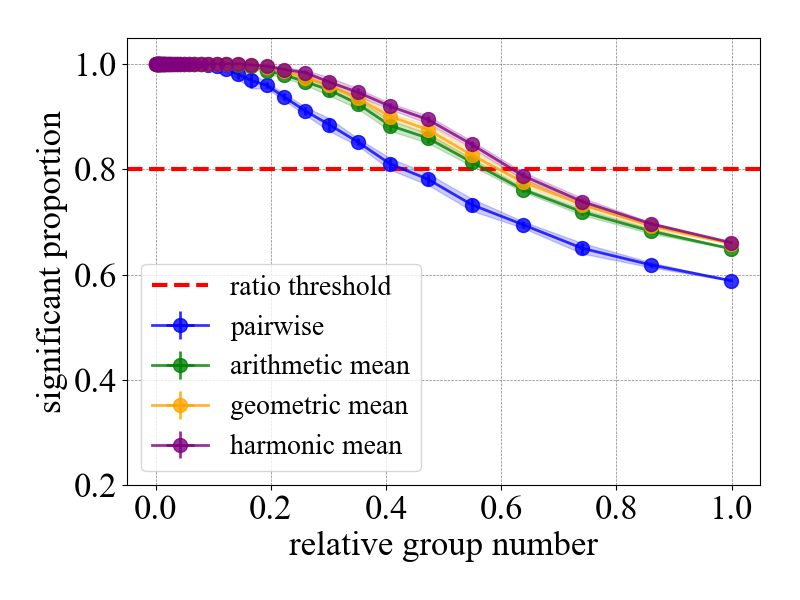}\\
{\footnotesize(a) Trivia QA dataset} & {\footnotesize(b) CNN daily mail dataset }& {\footnotesize(c) Bird dataset}
\end{tabular}
 \caption{Verification of hypotheses using the fraction of verified groups $\rho (n_g)$ as a function of the relative number of groups $n_g/n$ with Jaccard pairwise similarity and three different aggregation functions on $3$ representative datasets for QA, summarization, and text-to-SQL tasks. Complete results for $8$ datasets are included in Figure \ref{fig:agg_SQL} in the Appendix. } 
\label{fig:agg_main_paper} 
\vspace{-0.4cm}
\end{figure*}

\noindent\textbf{QA Task.} Following prior work on UQ~\citep{kuhn2022semantic,lin2023generating}, we use the open-book conversational question answering dataset CoQA \citep{reddy2019coqa}, the closed-book QA dataset TriviaQA \citep{joshi2017triviaqa}, as well as the more challenging closed-book QA dataset Natural Questions (NQ) \citep{kwiatkowski2019nq}. We consider the corresponding dev sets, consisting of 7983, 9960, and 3610 questions for CoQA, TriviaQA, and NQ respectively. We generate responses using \textbf{Granite 13B} \citep{mishra2024granite} (default) and \textbf{LLaMA 2 70B} \citep{touvron2023llama}.


\noindent\textbf{Summarization Task.} We use the CNN DailyMail Version 3.0.0~\citep{see-etal-2017-get,HermannKGEKSB15} and Extreme Summarization (XSum)~\citep{xsum-emnlp} datasets. For each dataset, we generate summaries using \textbf{LLaMA 3 8B} \citep{MetaLlama3} (default)  and \textbf{Mistral 8x7B} \citep{mistral8x7b} for the first 1000 documents.

\noindent\textbf{Text-to-SQL Task.}
We consider various real-world benchmark datasets for the task of converting natural language queries to SQL.
Spider~\citep{yu2018spider} is a popular benchmark that covers 138 domains, 
and Spider-Realistic~\citep{deng2021structure} is a more challenging version that modifies the natural language queries 
to avoid explicit mention of column names. 
BIRD~\citep{li2024can} is a recent cross-domain benchmark of 95 databases covering more than 37 professional domains. We use the dev sets for all 3 datasets, which include 1034, 508, and 1533 queries respectively.
We generate SQL with \textbf{Codellama 34 B}~\citep{rozière2024code} (default) and a \textbf{Deepseek 33 B} model~\citep{guo2024deepseekcoder} that is further fine-tuned with LoRA~\citep{hu2021lora} using Spider's training  dataset.

\begin{figure*}
\setlength{\tabcolsep}{-0.05cm}
\setlength{\arrayrulewidth}{0.1cm} 
\begin{tabular}{ccc}
\includegraphics[width=.35\textwidth]{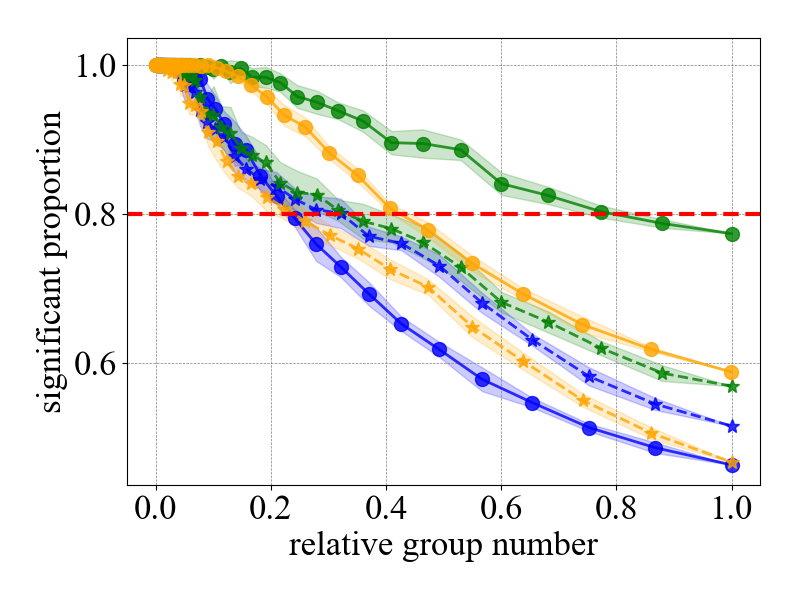}&
\includegraphics[width=.35\textwidth]{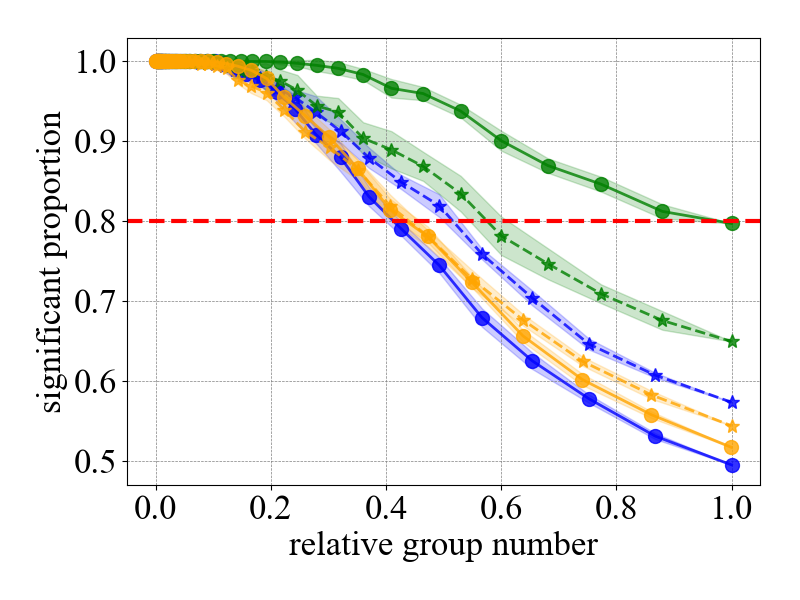}&
\includegraphics[width=.35\textwidth]{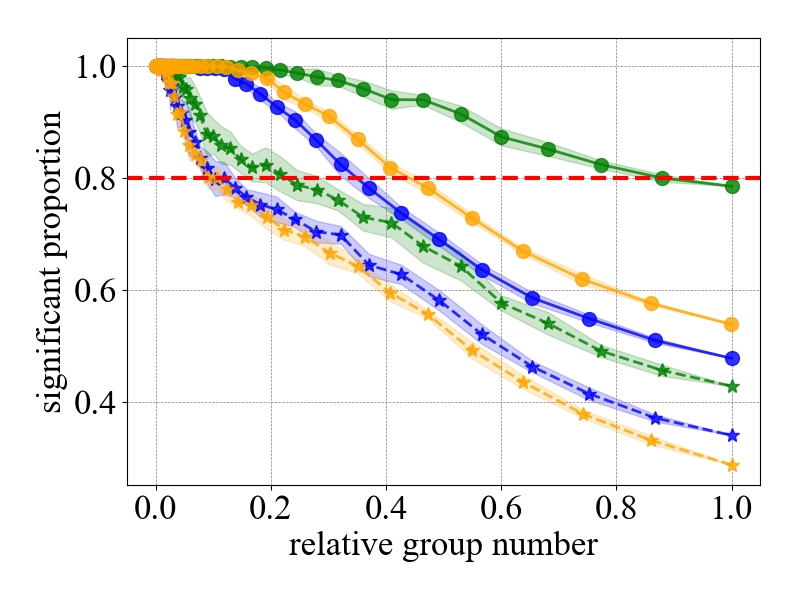}\\
\multicolumn{3}{c}{\includegraphics[width=0.9\textwidth]{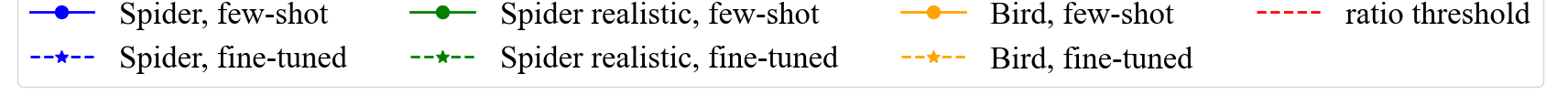}}\\
{\footnotesize(a) Hypo 1 (Sim-Any)} & {\footnotesize(b) Hypo 2 (Sim-Correct)}& {\footnotesize(c) Hypo 3 (Sim-Separate)}
\end{tabular}
\vspace{-0.3cm}
\caption{Impact of different datasets and models on verifying Hypothesis 1--3 for the text-to-SQL task. }
\label{fig:dataset_model_SQL} 
\vspace{-0.4cm}
\end{figure*}

\noindent\textbf{Other Experimental Details.}
We produce $5$ samples each over $6$ temperatures (from $0.25$ to $1.5$ in increments of $0.25$) from all LLMs, so as to obtain sufficient variability across generations~\citep{zhu2024hot}. An exploration of other sampling approaches and their ablation is provided in the Appendix. Most experiments use the Jaccard coefficient as similarity metric $s(\cdot)$, but we also consider variations of ROUGE metrics such as Rouge-1 and Rouge-L, and the cosine similarity between sentence BERT \url{https://sbert.net/} (sbert) representations of the generations. 
Wherever relevant, we consider the arithmetic mean of similarities as the aggregation function $f(\cdot)$, but we also consider simple extensions such as the geometric and harmonic mean.
For QA and summarization, a generated response is deemed correct if the Rouge-L score with respect to ground truth is no less than 0.5 and 0.2 respectively; Rouge-L has been used in this fashion for QA in prior work~\citep{kuhn2022semantic,lin2023generating}. Our results are robust with respect to Rouge-L thresholds that define correctness of generations, as noted by an ablation study in Appendix A.5. For text-to-SQL, a generated SQL is correct if it returns the same result as the ground truth SQL upon query execution on the underlying database.
We run Algorithm~\ref{alg: fraction_group_num} with $R=10$ repetitions.

\vspace{-0.2cm}
\subsection{Pairwise Consistency Statements}
\vspace{-0.2cm}

We begin by verifying the pairwise consistency hypotheses 1--3 (denoted H1--H3) using the Jaccard similarity measure across various datasets on the three tasks. 


We show the mean difference of correct similarity set $S^C$ and incorrect set $S^I$ in Figure \ref{fig:hypo_mean_diff} across diverse datasets and tasks. All mean differences are positive, indicating that all hypotheses are validated to some extent, but it is largest for H2. 
Recall that H2 suggests that compared to incorrect generations, correct generations are more similar to other correct generations. The validation of H3 indicates that incorrect generations have higher variability and are more diversely distributed than the correct generations. 
Note that in practical scenarios, one does not know the correct and incorrect labels in advance. 
Therefore, despite the stronger validation of H2 and H3, 
H1 serves as a practical approximation and its validation provides guidance for confidence estimation of each generation using similarity with respect to all other generations. 

In Figure \ref{fig:hypo_jaccard_pairwise_main_paper}, we illustrate the verification of all hypotheses at the group level using a plot of significant proportion (a.k.a. the fraction of verified groups) vs. the relative number of groups. We showcase only one representative dataset for each task in Figure \ref{fig:hypo_jaccard_pairwise_main_paper} and defer plots for other datasets to Figure  \ref{fig:hypo_jaccard_pairwise} in the Appendix. Conclusions drawn regarding the correctness of the hypotheses at the dataset level remain applicable at the group level. 
We also provide similar plots for the hypotheses using Rouge-L similarity in Figure \ref{fig:hypo_rougeL_pairwise} in the Appendix. The trend is the same as those using Jaccard.

\vspace{-0.1cm}
\subsection{Aggregated Consistency Statements}
\vspace{-0.1cm}
Next we investigate the correctness of the aggregated consistency statements and impact of different aggregation functions across diverse datasets on three tasks: QA, summarization, and text-to-SQL. We provide the verification plots of the aggregated version of three hypotheses in Figure \ref{fig:hypo_SQL_agg_arith} in the Appendix. All hypotheses are again verified to some extent. 
We choose H1 to study the effect of different aggregation functions versus the pairwise similarity statement due to its practical implications for UQ. 

We show results for representative datasets in Figure \ref{fig:agg_main_paper} and defer others to Figure \ref{fig:agg_SQL} in the Appendix. The conclusions for QA and summarization tasks are identical -- since there is less data available for the aggregated consistency statement than the pairwise one, the aggregated hypothesis tends to be statistically less significant. However, this seems to flip for the text-to-SQL task, 
especially for the harmonic mean. 
This suggests that aggregated consistency statements may be particularly effective in summarizing useful information from pairwise similarities, at least for some tasks.


\begin{table*}
\centering
\resizebox{\linewidth}{!}{%
\begin{tabular}{llcccccl}\toprule
Dataset (Model) & & \multicolumn{2}{c}{CoQA (Granite)} & \multicolumn{2}{c}{Spider (Codellama)} & \multicolumn{2}{c}{TriviaQA (Granite)} \\\cmidrule(lr){3-4}\cmidrule(lr){5-6}\cmidrule(lr){7-8} & & AUROC $\uparrow$ & AUARC $\uparrow$    & AUROC $\uparrow$ &  AUARC $\uparrow$ & AUROC $\uparrow$  &  AUARC $\uparrow$ \\\midrule
Baselines & always 1   
& $0.5 {\scriptstyle\pm 0.0}$ & $0.62 {\scriptstyle\pm 0.003}$ & $0.5 {\scriptstyle\pm 0.0}$ & $0.21 {\scriptstyle\pm 0.009}$ & $0.5 {\scriptstyle \pm 0.0}$ & $0.35 {\scriptstyle\pm 0.005}$ \\
 & avg. prob  & $0.75 {\scriptstyle\pm 0.004}$ & $0.79 {\scriptstyle\pm 0.006}$ & $0.62 {\scriptstyle\pm 0.012}$  & $0.26 {\scriptstyle\pm 0.015}$  & $0.74 {\scriptstyle\pm 0.002}$ & $0.51 {\scriptstyle\pm 0.005}$ \\
 & spec-ecc  & $0.18 {\scriptstyle\pm 0.008}$ & $0.37 {\scriptstyle\pm 0.003}$ & $0.31 {\scriptstyle\pm 0.008}$  & $0.13 {\scriptstyle\pm 0.006}$  & $0.20 {\scriptstyle\pm 0.002}$ & $0.17 {\scriptstyle\pm 0.002}$ \\
 & p(True)   & $0.59 {\scriptstyle\pm 0.026}$ & $0.70 {\scriptstyle\pm 0.015}$ & $0.54 {\scriptstyle\pm 0.009}$ & $0.23 {\scriptstyle\pm 0.011}$ & $0.68 {\scriptstyle\pm 0.005}$ & $0.47 {\scriptstyle\pm 0.007}$ \\ 
  \midrule
  Black-box & arith-agg   & $0.82 {\scriptstyle\pm 0.006}$ & $0.82 {\scriptstyle\pm 0.007}$ & $0.74 {\scriptstyle\pm 0.006}$  & $0.34 {\scriptstyle\pm 0.013}$  & $\textbf{0.80} {\scriptstyle\pm 0.002}$ & $\textbf{0.57} {\scriptstyle\pm 0.004}$ \\
 (Agg. sims) & geom-agg   & $\textbf{0.85} {\scriptstyle\pm 0.008}$ & $\textbf{0.83} {\scriptstyle\pm 0.007}$ & $\textbf{0.76} {\scriptstyle\pm 0.011}$  & $\textbf{0.35} {\scriptstyle\pm 0.017}$  & $\textbf{0.80} {\scriptstyle\pm 0.003}$ & $\textbf{0.57} {\scriptstyle\pm 0.004}$ \\
  & harm-agg   & $\textbf{0.85} {\scriptstyle\pm 0.008}$ & $\textbf{0.83} {\scriptstyle\pm 0.008}$ & $\textbf{0.76} {\scriptstyle\pm 0.010}$  & $\textbf{0.35} {\scriptstyle\pm 0.016}$  & $\textbf{0.80} {\scriptstyle\pm 0.003}$ & $\textbf{0.57} {\scriptstyle\pm 0.004}$ \\
\bottomrule
\end{tabular}
}
\caption{Comparing different aggregation approaches with Jaccard similarities for black-box UQ. Error bars for AUROC and AUARC are computed over $5$ runs that randomly choose $50\%$ of the dataset for testing.  
}
\vspace{-0.3cm}
\label{table-bb-uq}
\end{table*}

\subsection{Ablation Study: Similarity Metric \& Aggregation Function}

In this section, we conduct an ablation study of similarity measures and aggregation functions for the verification of H1. The results for QA and text-to-SQL tasks in terms of both mean difference and maximum relative number of groups to achieve $\rho^*$ amount of verified group are shown in Tables \ref{table:sim_agg_study_natural_question} and \ref{table:sim_agg_study_spider}. Due to space limitations, we defer Table \ref{table:sim_agg_study_spider} to the Appendix. 
For similarity measures, we test Jaccard, Rouge-1, Rouge-L, and the cosine similarity between sbert representation for both tasks, and another metric that identifies the type of SQL output~\citep{pourreza2024din} for the text-to-SQL datasets. For  aggregation functions, we use the arithmetic, geometric and harmonic means. 

The key insights from these two tables are: 1) similar to the conclusion in Section 4.2, aggregation degrades the verification extent in the QA task, whereas it enhances validation of the hypothesis in the text-to-SQL task; 2) the hypothesis is more true under Rouge-1 and Rouge-L scores than Jaccard similarity across both tasks; 3) Jaccard, Rouge-1, Rouge-L are robust in both tasks; 4) sbert similarity occasionally outperforms others in the max. relative group number to achieve a certain fraction of verified groups; 
5) SQL output type similarity is least reliable for verifying the hypothesis in the text-to-SQL task. Furthermore, since Jaccard similarity is the most computationally efficient measure besides being the most robust, we use it for most experiments.

\vspace{-0.3cm}
\subsection{Ablation Study: Datasets \& Models}
\vspace{-0.2cm}

The correctness of  a hypothesis is a property of both datasets and models. Here we investigate the effect of different datasets and models on the validation of pairwise statements for the text-to-SQL task as shown in Figure \ref{fig:dataset_model_SQL}. We refer the Codellama and the fine-tuned Deepseek model as the  few-shot and fine-tuned model, respectively. The Spider Realistic dataset is one where the consistency hypothesis holds to the greatest extent, followed by Bird and then Spider. We observe that all the hypotheses are stronger for few-shot models as compared to fine-tuned models, for both the Spider Realistic and Bird datasets. 
This may occur because incorrect instances may have less variability after fine-tuning the models for specific tasks. 

\subsection{Ablation Study: Sampling Method}
\begin{figure}
\includegraphics[width=0.5\textwidth]{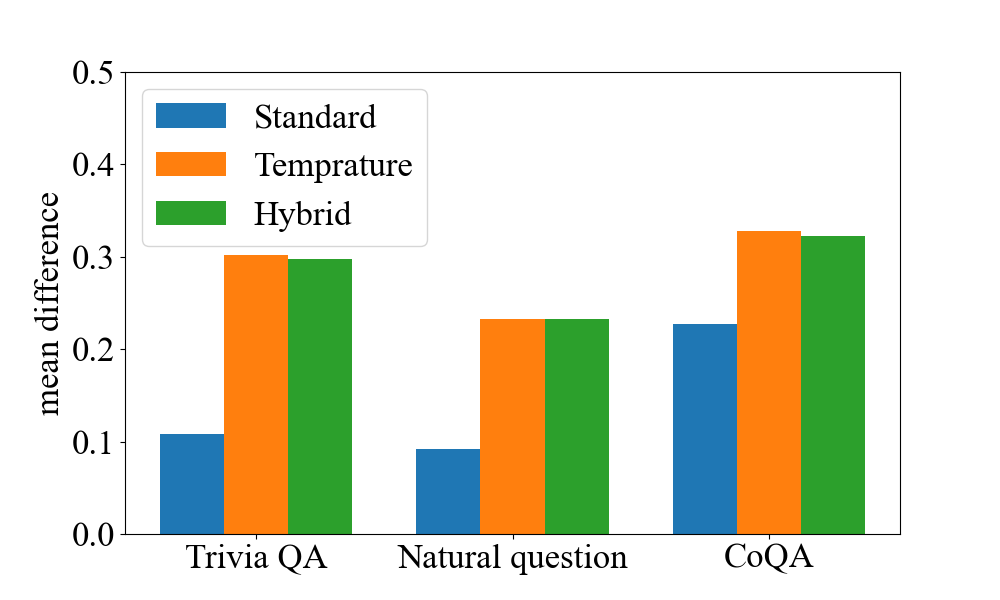}
\caption{Effects of sampling approaches to the verification of H1 on the QA task with Granite model. }
\label{fig:sampling}
\vspace{-0.2cm}
\end{figure}

We evaluate the consistency hypothesis for the QA task using three sampling methods: 1) standard sampling, where tokens are generated through decoding at a fixed temperature of $0.25$, 2) temperature sampling, which involves generating one sample at varying temperatures, and 3) hybrid sampling, where multiple samples are generated from multiple temperatures (this is what we used in other experiments). Figure \ref{fig:sampling} displays the mean difference between correct and incorrect clusters at the dataset level. Sampling at different temperatures (temperature and hybrid sampling) increases the variability of generations, thereby enhancing the validity of the consistency hypotheses. 

\section{
Confidence Estimation by Aggregation}

To demonstrate how rigorous statistical validation of the consistency hypothesis can guide UQ method design, we introduce a new UQ method based on the `Sim-Any' hypothesis (H1) as an illustrative example.

Given that the consistency hypothesis H1 is generally verified to various degrees across datasets and models in Section \ref{sec:empirical}, we investigate the use of simple aggregation methods including arithmetric, geometric and harmonic mean as a way of estimating confidence. Specifically, we treat the aggregated similarity of generation $j$ w.r.t other generations $\bar{s}_i^j = f(s_i^{j,1}, \cdots, s_i^{j,m})$ as its confidence. 
Note that the `degree' approach used after spectral clustering, as discussed in~\citep{lin2023generating}, corresponds to what we describe here as arithmetic mean aggregation. In contrast, the geometric and harmonic means are novel aggregation methods introduced in this work, inspired by their superior empirical performance on verifying consistency hypothesis in Table \ref{table:sim_agg_study_natural_question}. 
Both methods yield more conservative confidence estimates, typically lower than those produced by the arithmetic mean.

We evaluate confidence using two standard metrics. The \emph{Area Under the Receiver Operating Characteristic} (AUROC) evaluates confidence when used as a probabilistic classifier for  correctness, whereas
the \emph{Area Under the Accuracy Rejection Curve} (AUARC) is suitable when confidences are used for selective generation, i.e. for rejecting a fraction of the instances that one is least confident about. 

Table~\ref{table-bb-uq} compares various UQ approaches over 2 datasets for QA (CoQA, TriviaQA) and 1 for text-to-SQL (Spider). 
We consider $4$ baselines: a naive one that always returns a score of $1$, a white-box approach that uses the avg. log probability of tokens, a spectral clustering technique using `eccentricity'~\citep{lin2023generating}, and a white-box verbalized approach that prompts an LLM to determine whether a generation is correct and uses the logit of the answer (True/False)~\citep{kadavath2022language}.
Results suggest that the black-box UQ methods with the geometric and harmonic means outperform the baselines. This highlights the practical significance of our work around verifying the consistency hypothesis and demonstrates that the extent of verification can inform the design of effective new UQ methods.

\section{Conclusion}

In this paper, we proposed three mathematical formulations to formalize a common conjecture about consistency in LLM generations. We substantiated these hypotheses through a comprehensive empirical investigation on eight benchmark datasets (CoQA, TriviaQA, Natural Questions, CNN DailyMail, XSum, Spider, Spider-Realistic, and BIRD) spanning three tasks (QA, Summarization and Text-to-SQL),  establishing justification for consistency-based UQ methods. We conducted a thorough ablation study, analyzing factors such as aggregation methods, sampling techniques, datasets, and models to validate the consistency hypotheses. Inspired by the empirical findings, we also proposed a black box UQ method with new aggregation approaches over similarities and demonstrated its superior performance over some baselines, thereby highlighting how our extensive empirical study on consistency can inform practice.  
An investigation of other aggregation methods as well as a broader empirical study with tasks involving additional complexities are potential avenues for future work.

\section*{Acknowledgments}

This work was supported by IBM through the IBM-Rensselaer Future of Computing Research Collaboration.


\bibliography{refs}

\appendix
\newpage

\onecolumn

\vspace{-1cm}

\title{The Consistency Hypothesis in Uncertainty Quantification \\ for Large Language Models (Supplementary Material)}

\maketitle


\doparttoc 
\faketableofcontents 


\part{} 
\parttoc 

\section{Additional Experimental Results}

\begin{figure*}[htb]
\setlength{\tabcolsep}{-0.05cm}
\begin{tabular}{cccc}
\includegraphics[width=.26\textwidth]{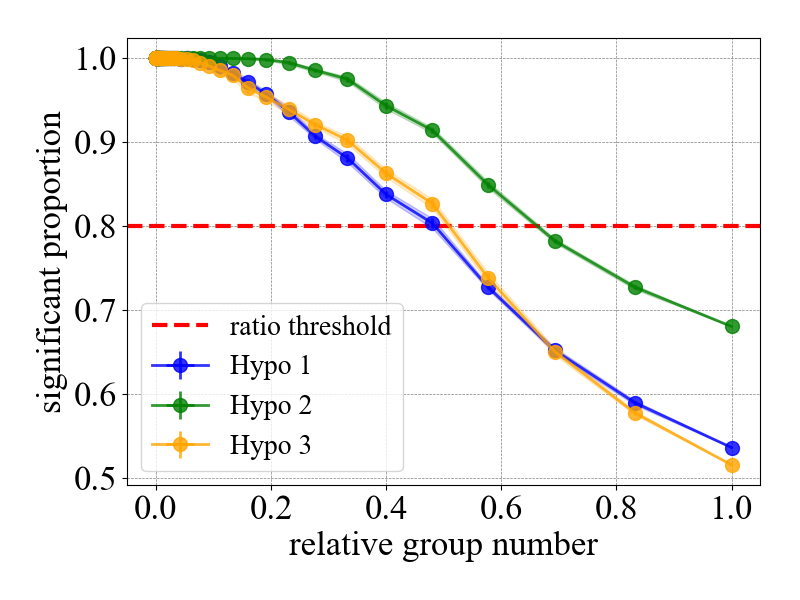} &
\includegraphics[width=.26\textwidth]{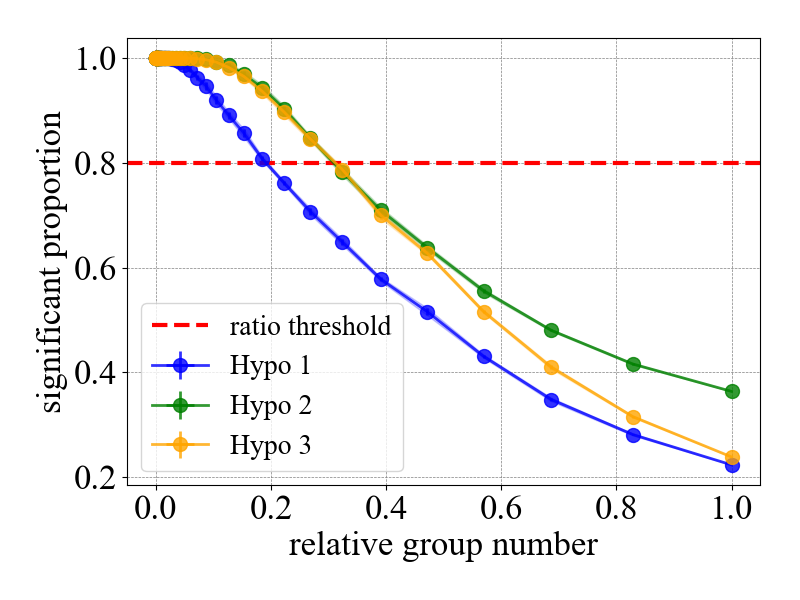} &
\includegraphics[width=.26\textwidth]{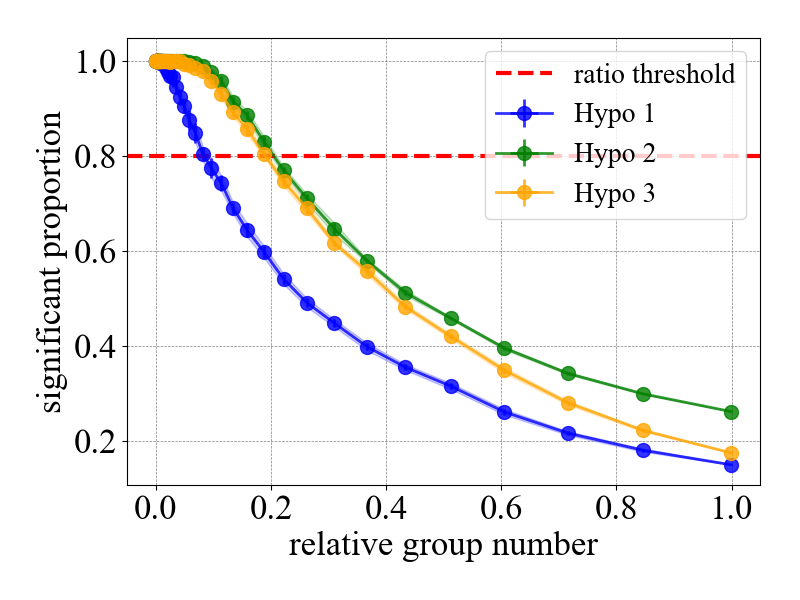} &
\includegraphics[width=.26\textwidth]{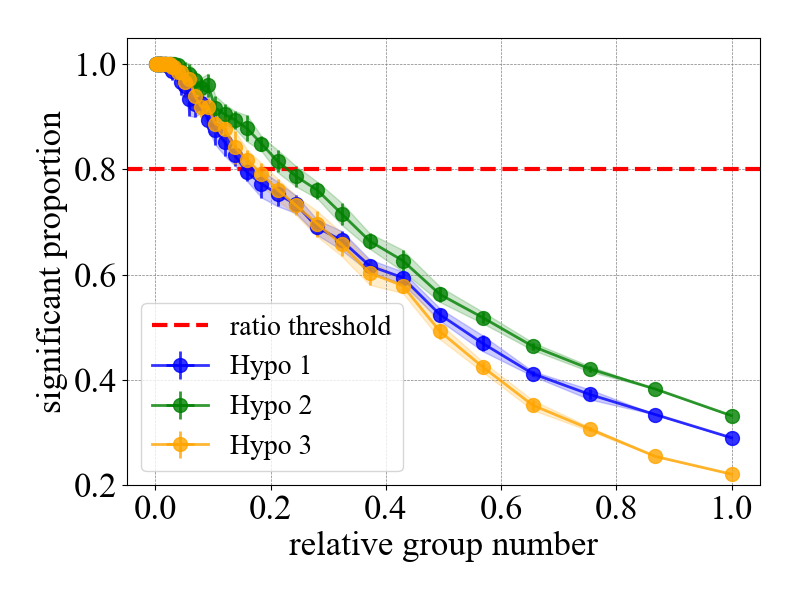} \\
{\footnotesize (a) CoQA dataset} & {\footnotesize (b) Trivia QA dataset} & {\footnotesize (c) Natural Question dataset} & {\footnotesize (d) CNN daily mail dataset} \\
\includegraphics[width=.26\textwidth]{figures/ratio_hypos_xsum_sim_jaccard.png} &
\includegraphics[width=.26\textwidth]{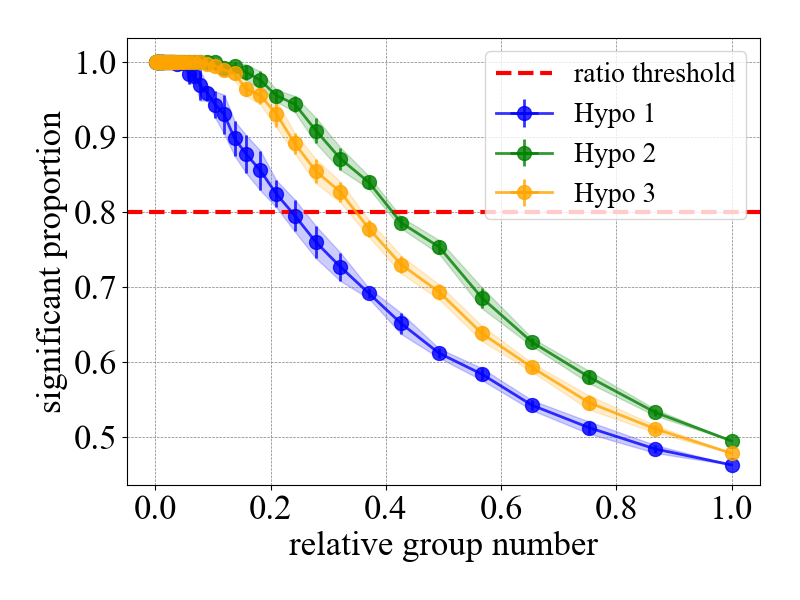} &
\includegraphics[width=.26\textwidth]{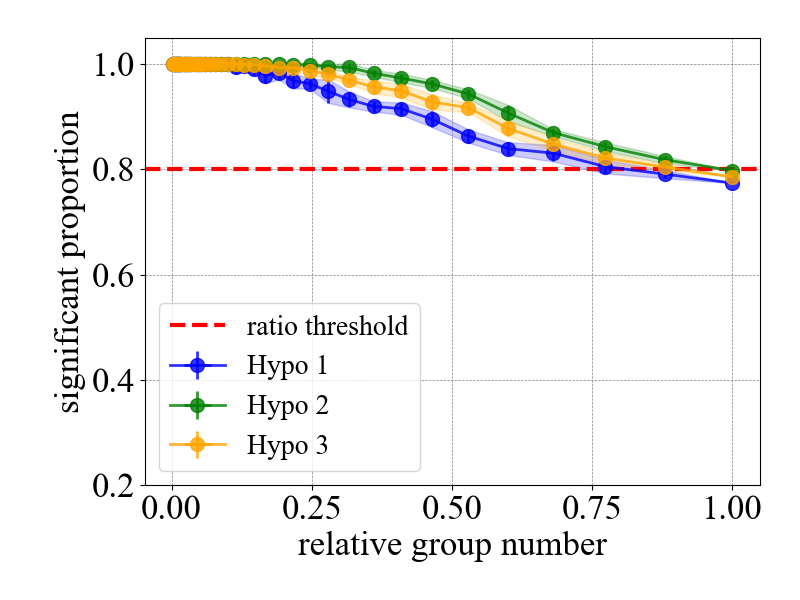} &
\includegraphics[width=.26\textwidth]{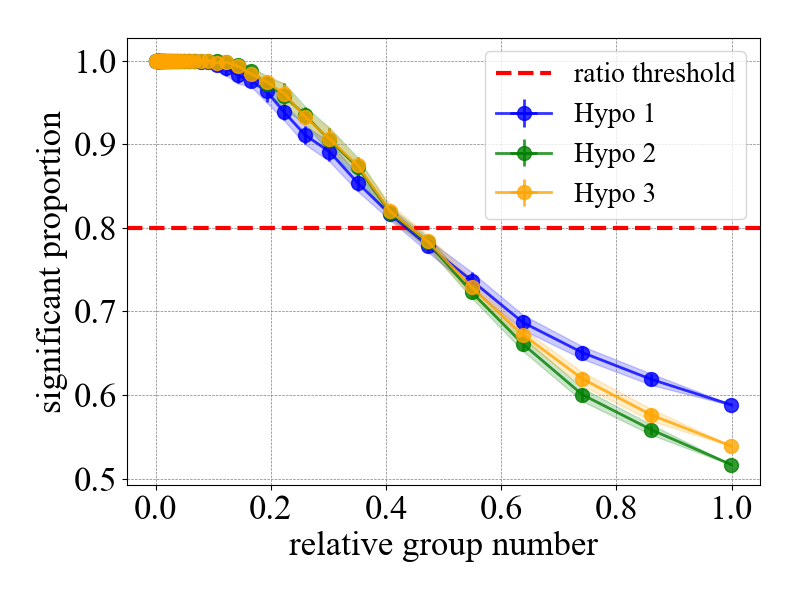} \\
{\footnotesize (e) XSum dataset} & {\footnotesize (f) Spider dataset} & {\footnotesize (g) Spider Realistic dataset} & {\footnotesize (h) Bird dataset} \\
\end{tabular}
\caption{Verification of three hypotheses using Jaccard pairwise similarity on various datasets for QA, summarization, and text-to-SQL tasks.}
\label{fig:hypo_jaccard_pairwise}
\end{figure*}

In this section, we include the omitted figures and tables from the main paper. 
\subsection{Verification of Pairwise Consistency Statements}
\begin{figure*}
\setlength{\tabcolsep}{-0.05cm}
\setlength{\arrayrulewidth}{0.01cm}
\begin{tabular}{cccc}
\includegraphics[width=.25\textwidth]{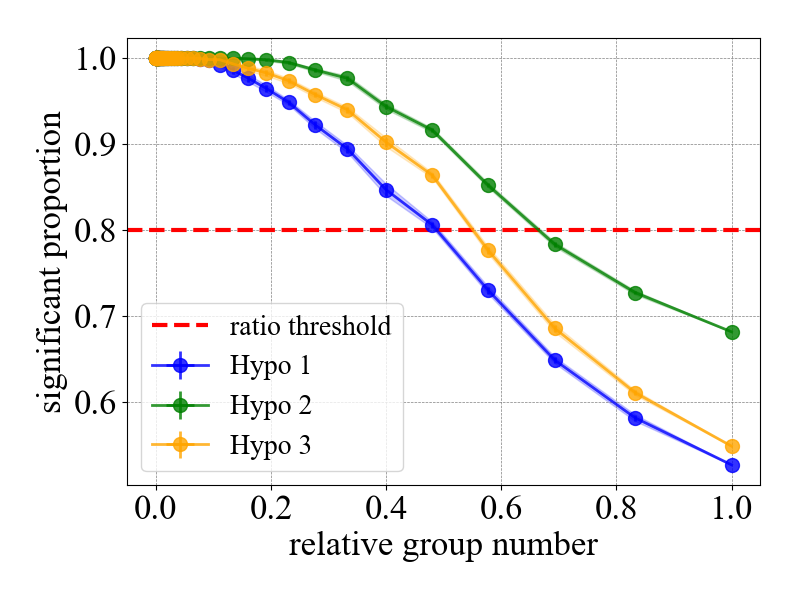}&
\includegraphics[width=.25\textwidth]{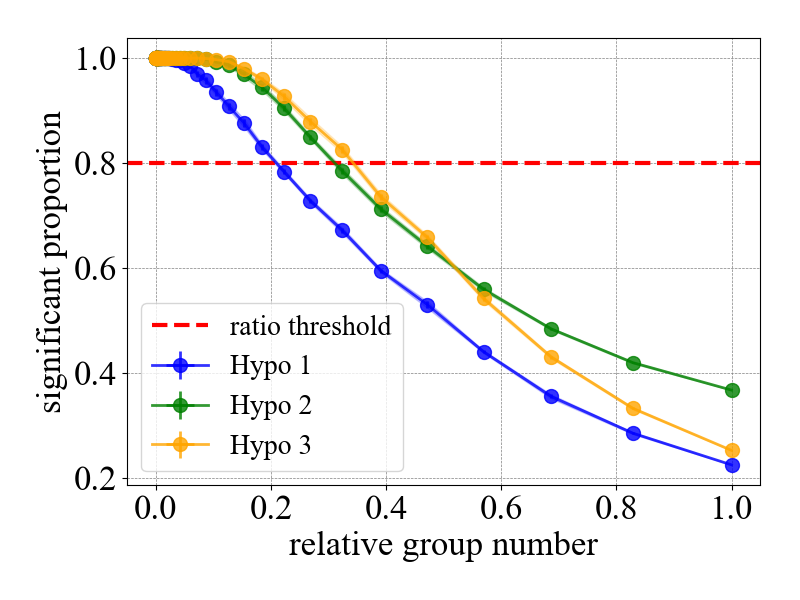}&
\includegraphics[width=.25\textwidth]{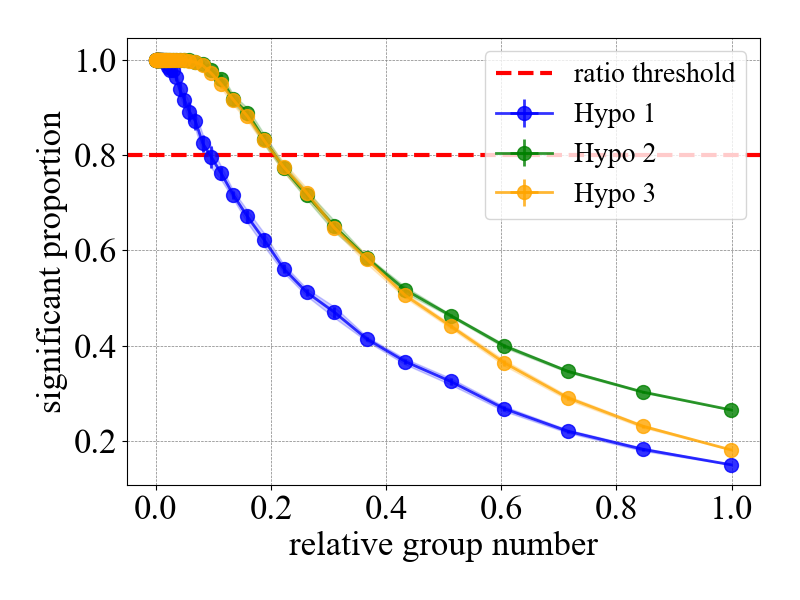}&
\includegraphics[width=.25\textwidth]{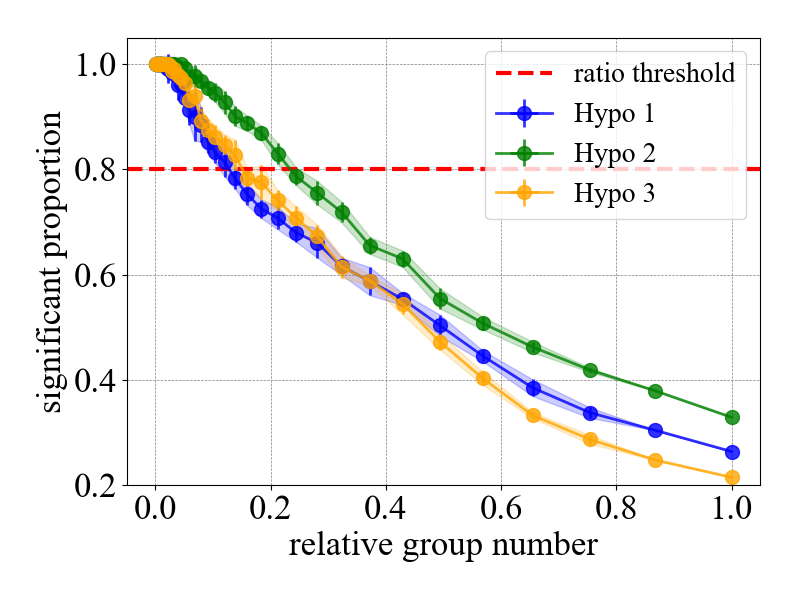}\\
{\footnotesize(a) CoQA dataset} & {\footnotesize(b) Trivia QA dataset}& {\footnotesize(c) Natural Question dataset}& {\footnotesize(d) CNN daily mail dataset}\\
\includegraphics[width=.25\textwidth]{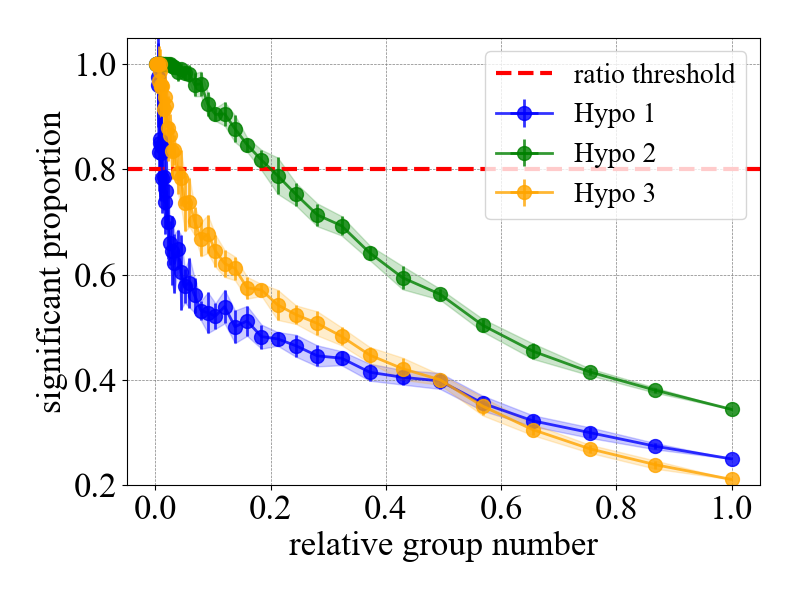}&
\includegraphics[width=.25\textwidth]{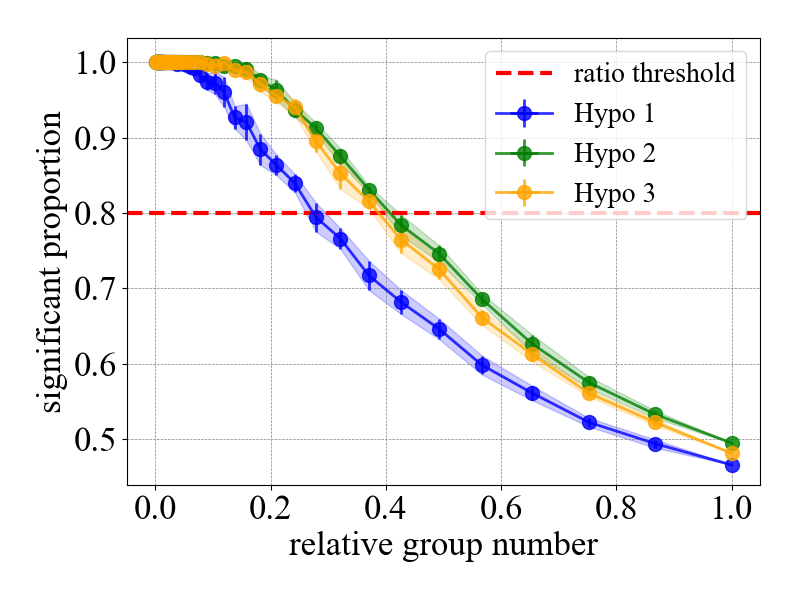}&
\includegraphics[width=.25\textwidth]{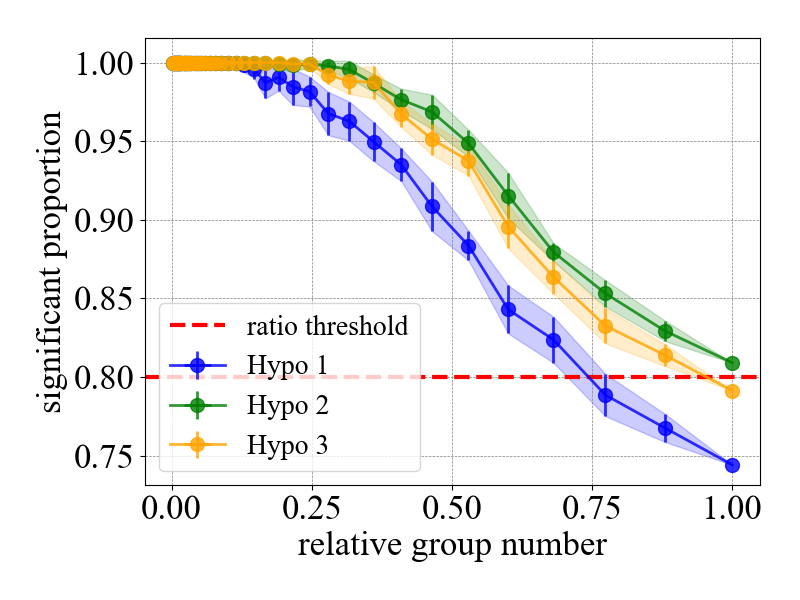}&
\includegraphics[width=.25\textwidth]{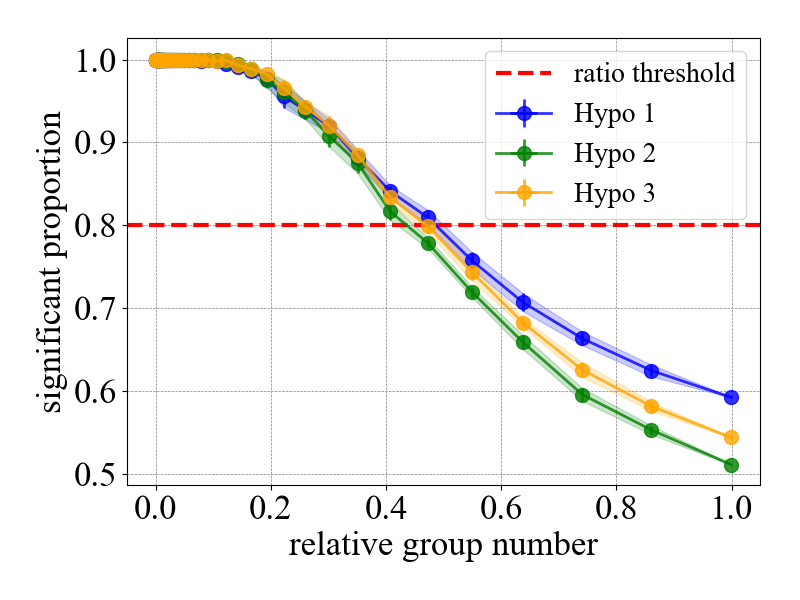}\\
{\footnotesize(e) XSum dataset}& {\footnotesize(f) Spider dataset} & {\footnotesize(g) Spider Realistic dataset}& {\footnotesize(h) Bird dataset}
\end{tabular}
 \caption{Verification of different hypotheses with Rouge-L pairwise similarity on QA, summarization, and text-to-SQL tasks for 8 datasets. }
\label{fig:hypo_rougeL_pairwise} 
\vspace{-0.4cm}
\end{figure*}

In this section, we provide the complete verification results for the pairwise consistency hypotheses for all the datasets in Figure \ref{fig:hypo_jaccard_pairwise}. We also provide the validation of the pairwise consistency hypotheses using Rouge-L similarity in Figure \ref{fig:hypo_rougeL_pairwise}. All of the hypotheses are verified to some extent across all of the datasets. Among those, consistency hypotheses are more true on the QA and text-to-SQL tasks than the summarization task. This may due to the higher viability of correct answers in the text summarization task.

\subsection{Verification of Aggregated Consistency Statements}
\begin{figure*}[htb]
\setlength{\tabcolsep}{-0.05cm}
\begin{tabular}{cccc}
\includegraphics[width=.25\textwidth]{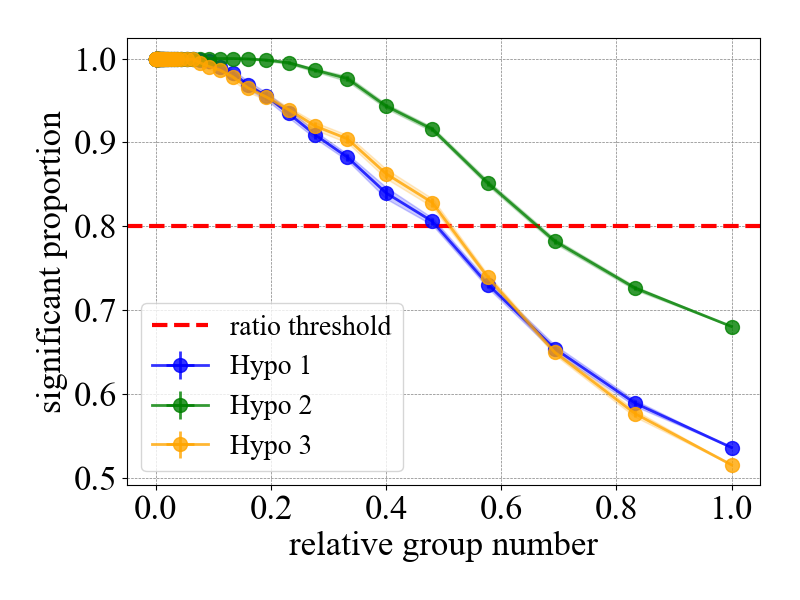}&
\includegraphics[width=.25\textwidth]{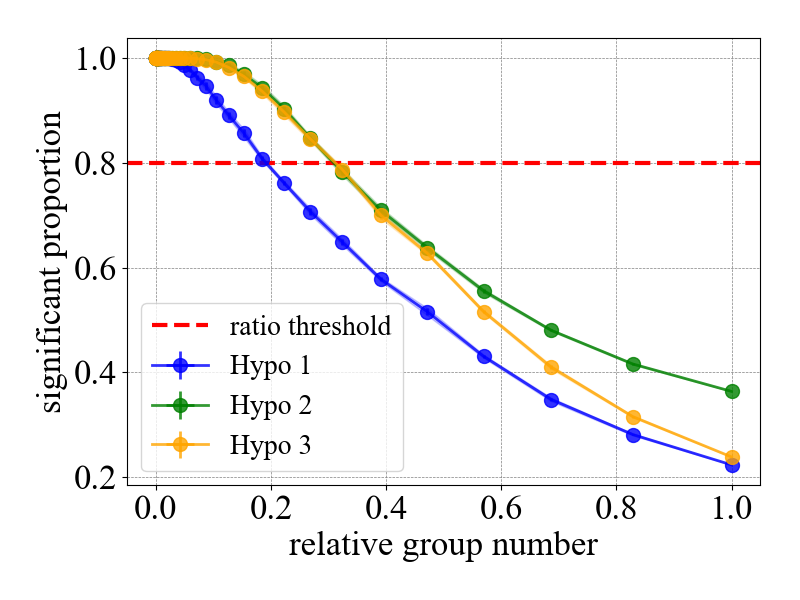}&
\includegraphics[width=.25\textwidth]{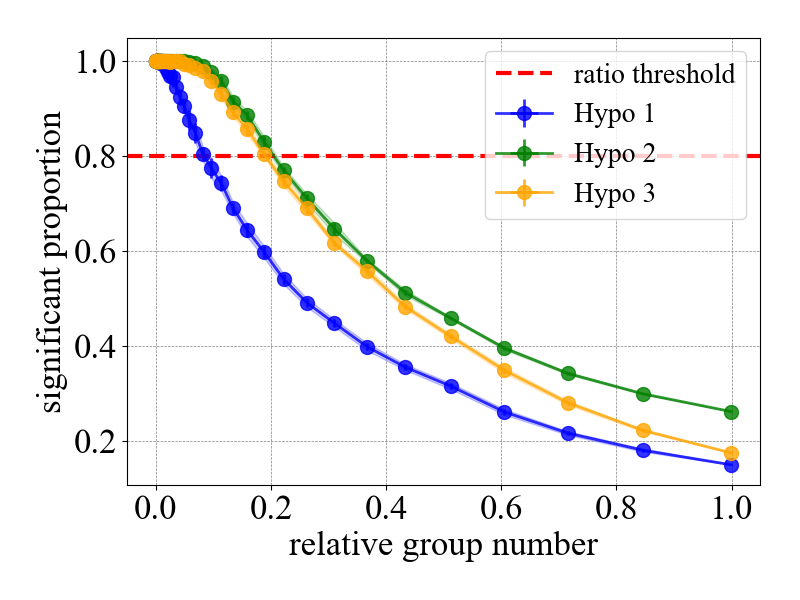}
&
\includegraphics[width=.25\textwidth]{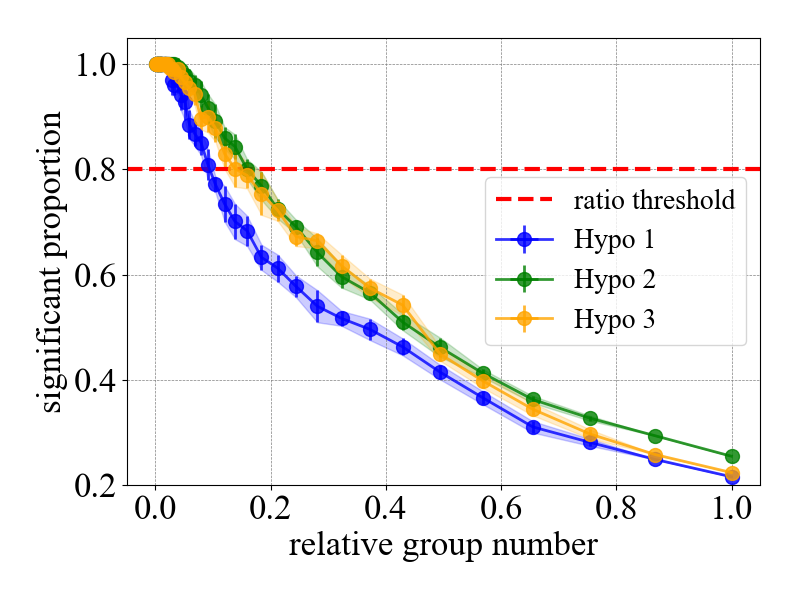}\\
{\footnotesize(a) CoQA dataset} & {\footnotesize(b) Trivia QA dataset}& {\footnotesize(c) Natural Question dataset}& {\footnotesize(d) CNN daily mail dataset}\\
\includegraphics[width=.25\textwidth]{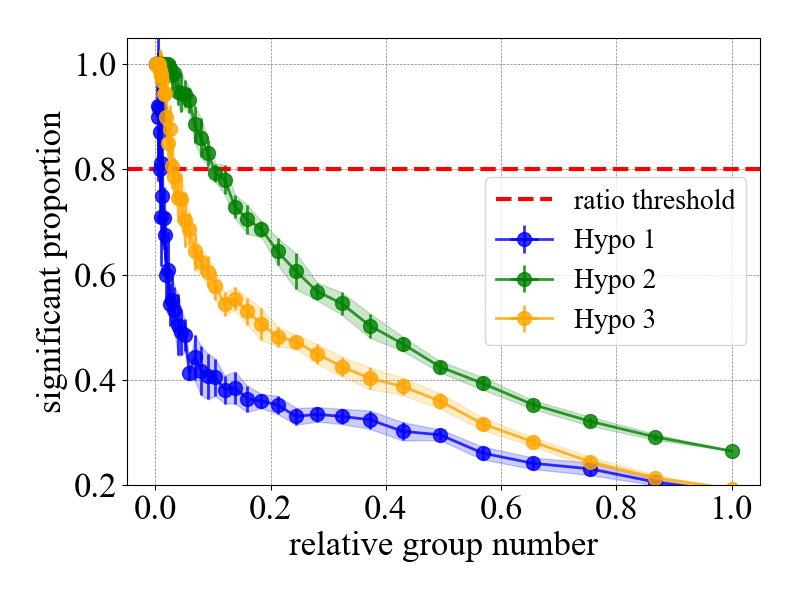}&
\includegraphics[width=.25\textwidth]{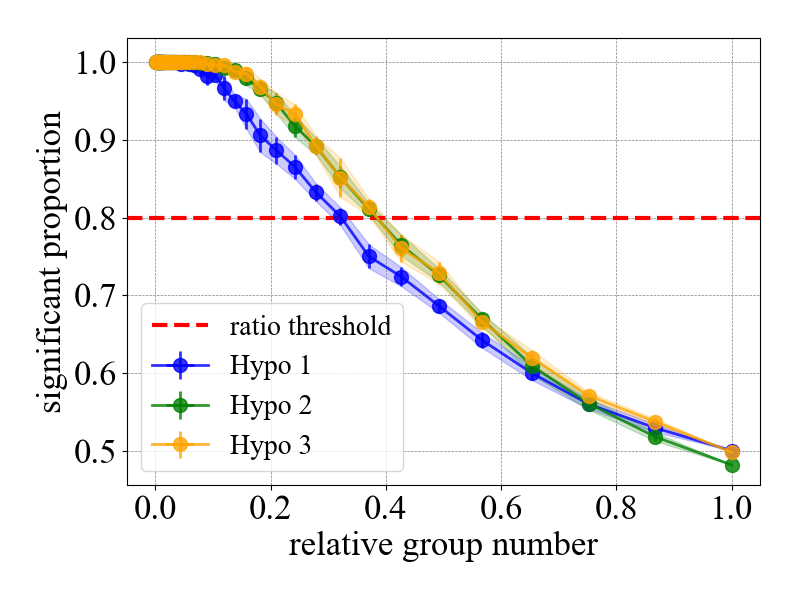}&
\includegraphics[width=.25\textwidth]{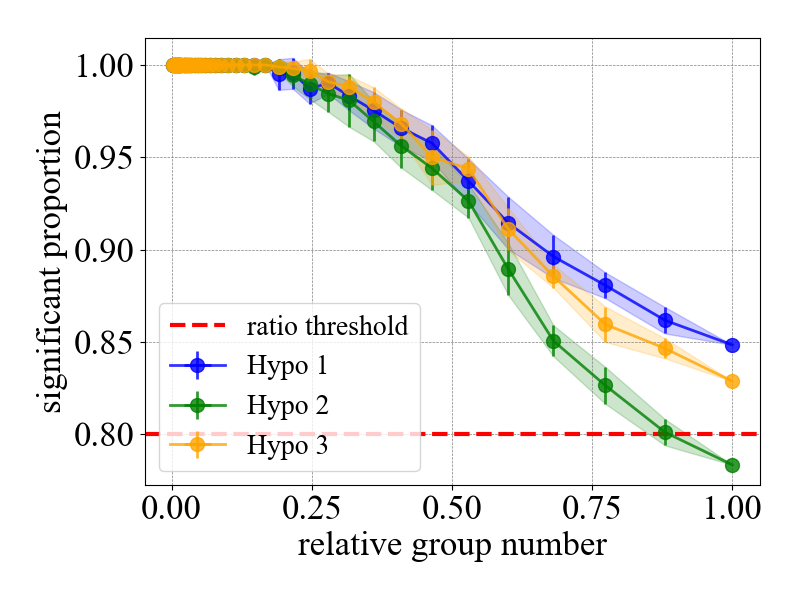}
&
\includegraphics[width=.25\textwidth]{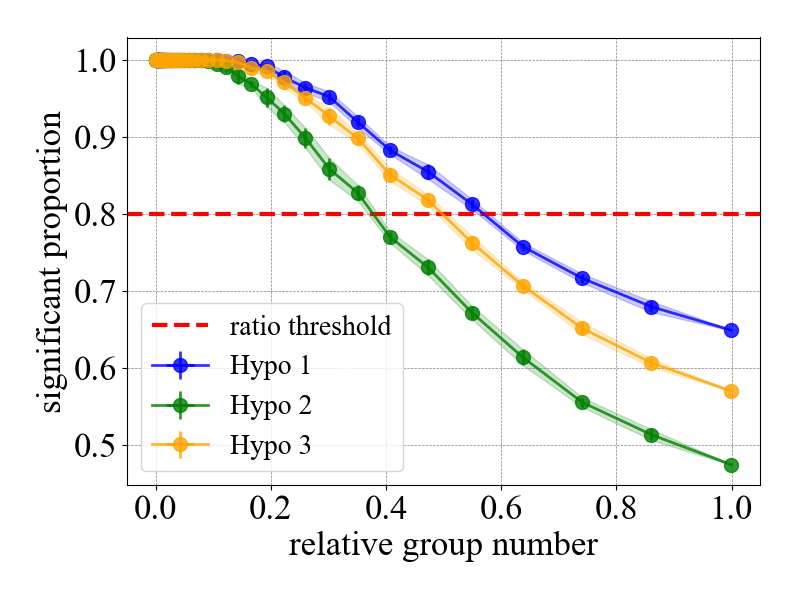}
\\
{\footnotesize(d) XSum dataset}&{\footnotesize(d) Spider dataset} & {\footnotesize(e) Spider Realistic dataset}& {\footnotesize(f) Bird dataset}
\end{tabular}
 \caption{Verification of different hypotheses using Jaccard  similarity and arithmetic mean as aggregation function on QA, summarization, and text-to-SQL tasks for 8 datasets. }
\label{fig:hypo_SQL_agg_arith} 
\vspace{-0.4cm}
\end{figure*}

\begin{figure*}[htb]
\setlength{\tabcolsep}{-0.05cm}
\begin{tabular}{cccc}
\includegraphics[width=.26\textwidth]{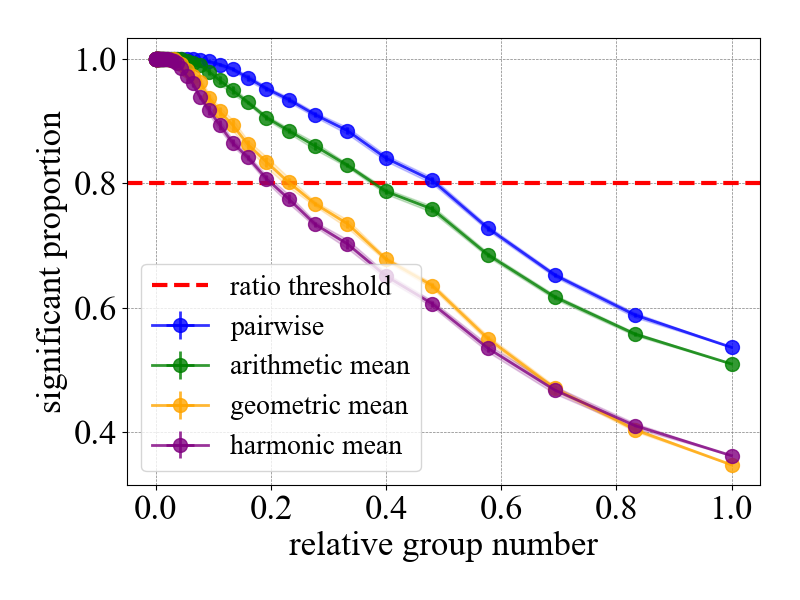} &
\includegraphics[width=.26\textwidth]{figures/ratio_aggregation_triviaqa_granite.png} &
\includegraphics[width=.26\textwidth]{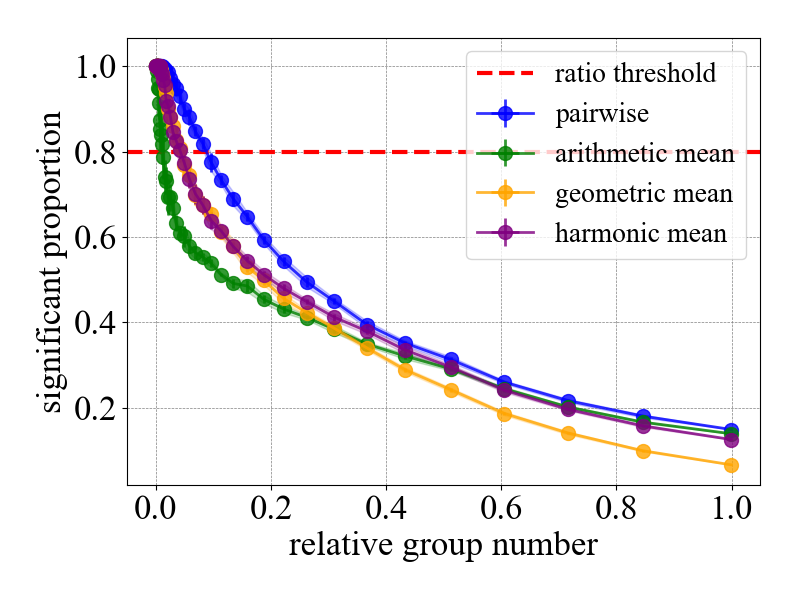} &
\includegraphics[width=.26\textwidth]{figures/ratio_aggregation_daily_mail.png} \\
{\footnotesize(a) CoQA dataset} & {\footnotesize(b) Trivia QA dataset} & {\footnotesize(c) Natural Question dataset} & {\footnotesize(d) CNN daily mail dataset} \\
\includegraphics[width=.26\textwidth]{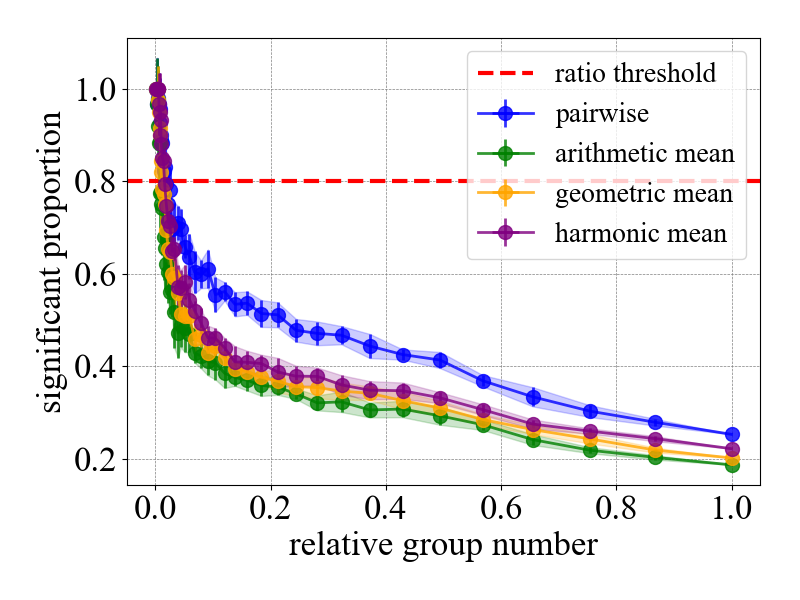} &
\includegraphics[width=.26\textwidth]{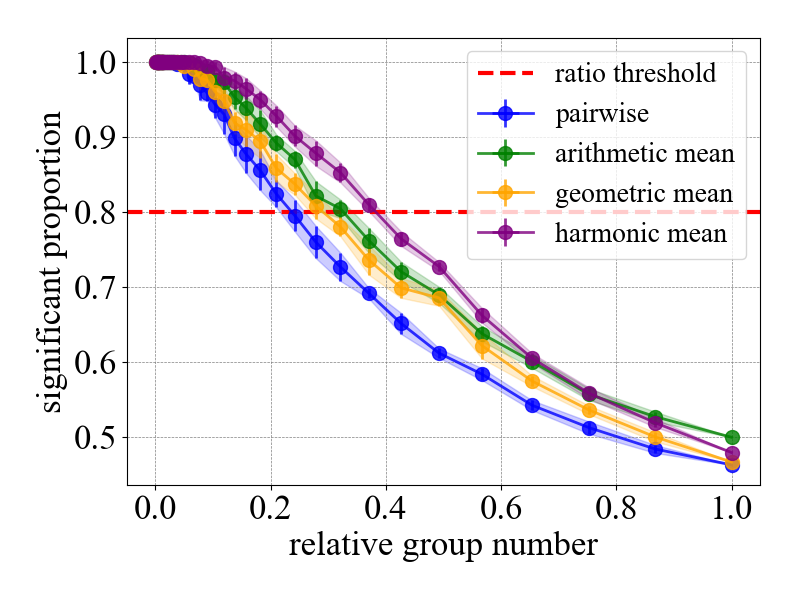} &
\includegraphics[width=.26\textwidth]{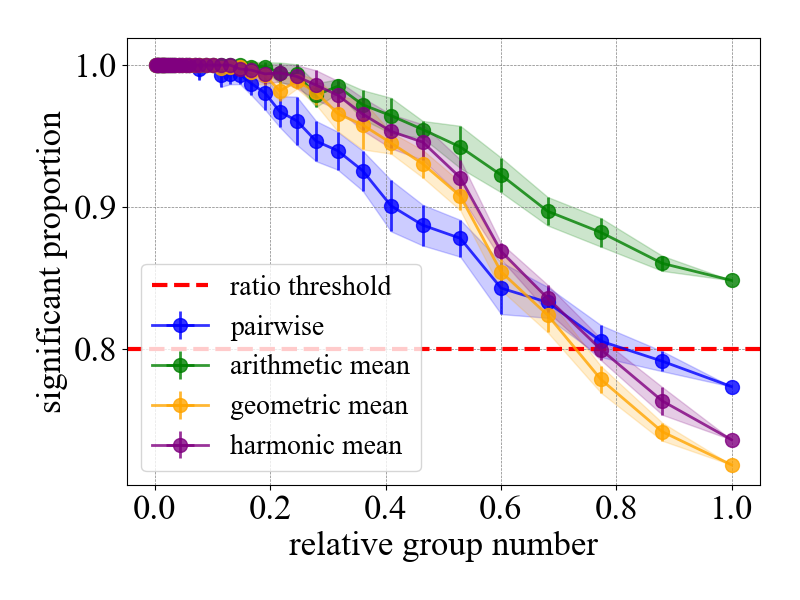} &
\includegraphics[width=.26\textwidth]{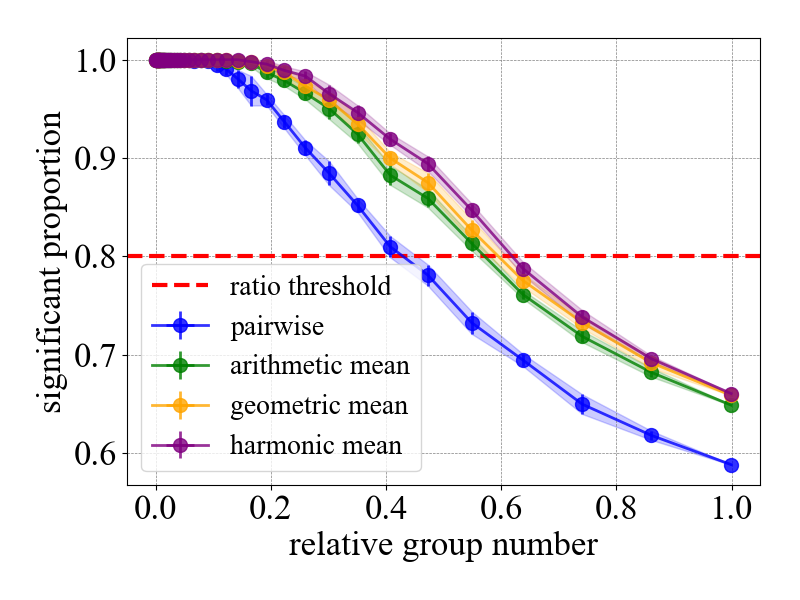} \\
{\footnotesize(e) XSum dataset} & {\footnotesize(f) Spider dataset} & {\footnotesize(g) Spider Realistic dataset} & {\footnotesize(h) Bird dataset} \\
\end{tabular}
\caption{Impact of aggregation functions for verifying aggregation version of Hypothesis 1 on different datasets in QA, summarization, and text-to-SQL tasks for 8 datasets.}
\label{fig:agg_SQL} 
\vspace{-0.2cm}
\end{figure*}

In this section, we provide the complete verification results for the aggregated consistency hypotheses with Jaccard similarity for all the datasets in Figure \ref{fig:hypo_SQL_agg_arith} and Figure \ref{fig:agg_SQL}. It can be seen that all three hypotheses are validated to some extent in all of the datasets. Surprisingly, while the aggregate consistency hypotheses generally exhibit lower validity compared to the pairwise version, especially in QA and summarization tasks due to reduced available similarity pairs, this trend does not hold for the text-to-SQL task. Notably, the harmonic mean aggregation achieves the highest validation performance, indicating its potential as an effective method for classifying the generations based on consistency.

\begin{table*}[htb]
\centering
\resizebox{\linewidth}{!}{%
\begin{tabular}{lcccccccccc}\toprule
\multirow{2}{*}{\shortstack{Aggregation/\\Similarity}} & \multicolumn{5}{c}{Mean difference when $n_g=10$ $\uparrow$} & \multicolumn{5}{c}{Max relative group number when $\rho^*=0.8$ $\uparrow$} \\
\cmidrule(lr){2-6} \cmidrule(lr){7-11} 
& Jaccard & Rouge-1 & Rouge-L & Sbert & SQL & Jaccard & Rouge-1 & Rouge-L & Sbert & SQL \\\midrule
\shortstack{pairwise \\ \vspace{4pt}}   
& \shortstack{$0.1519$ \\ \vspace{4pt} ${\scriptstyle \pm 0.0004}$} & \shortstack{$0.1395$ \\ \vspace{4pt} ${\scriptstyle \pm 0.0002}$} & \shortstack{$0.1482$ \\ \vspace{4pt} ${\scriptstyle \pm 0.0003}$} & \shortstack{$0.0815$ \\ \vspace{4pt} ${\scriptstyle \pm 0.0001}$} & \shortstack{$0.099$ \\ \vspace{4pt} ${\scriptstyle \pm 0.0003}$} & \shortstack{$0.2472$ \\ \vspace{4pt} ${\scriptstyle \pm 0.0003}$} & \shortstack{$0.2820$ \\ \vspace{4pt} ${\scriptstyle \pm 0.0001}$} & \shortstack{$0.2728$ \\ \vspace{4pt} ${\scriptstyle \pm 0.0004}$} & \shortstack{$0.2847$ \\ \vspace{4pt} ${\scriptstyle \pm 0.0002}$} & \shortstack{$0.1269$ \\ \vspace{4pt} ${\scriptstyle \pm 0.0002}$} \\
\shortstack{arithmetic \\ \vspace{4pt}}   
& \shortstack{$0.1625$ \\ \vspace{4pt} ${\scriptstyle \pm 0.0003}$} & \shortstack{$0.1964$ \\ \vspace{4pt} ${\scriptstyle \pm 0.0002}$} & \shortstack{$0.1956$ \\ \vspace{4pt} ${\scriptstyle \pm 0.0002}$} & \shortstack{$0.1410$ \\ \vspace{4pt} ${\scriptstyle \pm 0.0001}$} & \shortstack{$0.0656$ \\ \vspace{4pt} ${\scriptstyle \pm 0.0005}$} & \shortstack{$0.3261$ \\ \vspace{4pt} ${\scriptstyle \pm 0.0003}$} & \shortstack{$0.3933$ \\ \vspace{4pt} ${\scriptstyle \pm 0.0001}$} & \shortstack{$0.3854$ \\ \vspace{4pt} ${\scriptstyle \pm 0.0002}$} & \shortstack{$\bf 0.4144$ \\ \vspace{4pt} ${\scriptstyle \bf \pm 0.0003}$} & \shortstack{$0.0361$ \\ \vspace{4pt} ${\scriptstyle \pm 0.0001}$} \\
\shortstack{geometric \\ \vspace{4pt}}   
& \shortstack{$0.1365$ \\ \vspace{4pt} ${\scriptstyle \pm 0.0003}$} & \shortstack{$0.2081$ \\ \vspace{4pt} ${\scriptstyle \pm 0.0008}$} & \shortstack{$0.2007$ \\ \vspace{4pt} ${\scriptstyle \pm 0.0005}$} & \shortstack{0.1653\\ \vspace{4pt} ${\scriptstyle \pm 0.0001}$ } & \shortstack{$0.2010$ \\ \vspace{4pt} ${\scriptstyle \pm 0.0013}$} & \shortstack{$0.2858$ \\ \vspace{4pt} ${\scriptstyle \pm 0.0003}$} & \shortstack{$0.2951$ \\ \vspace{4pt} ${\scriptstyle \pm 0.0004}$} & \shortstack{$0.2983$ \\ \vspace{4pt} ${\scriptstyle \pm 0.0003}$} & \shortstack{$0.0010$ \\ \vspace{4pt} ${\scriptstyle \pm 0.0001}$} & \shortstack{$0.1348$ \\ \vspace{4pt} ${\scriptstyle \pm 0.0001}$} \\
\shortstack{harmonic \\ \vspace{4pt}}   
& \shortstack{$0.1317$ \\ \vspace{4pt} ${\scriptstyle \pm 0.0003}$} & \shortstack{$\bf 0.2161$ \\ \vspace{4pt} ${\scriptstyle \bf \pm 0.0005}$} & \shortstack{$0.2058$ \\ \vspace{4pt} ${\scriptstyle \pm 0.0004}$} & \shortstack{$0.1928$ \\ \vspace{4pt} ${\scriptstyle \pm 0.0462}$} & \shortstack{$0.2012$ \\ \vspace{4pt} ${\scriptstyle \pm 0.0010}$} & \shortstack{$0.3710$ \\ \vspace{4pt} ${\scriptstyle \pm 0.0001}$} & \shortstack{$0.3558$ \\ \vspace{4pt} ${\scriptstyle \pm 0.0005}$} & \shortstack{$0.3648$ \\ \vspace{4pt} ${\scriptstyle \pm 0.0001}$} & \shortstack{$0.0034$ \\ \vspace{4pt} ${\scriptstyle \pm 0.0001}$} & \shortstack{$0.1386$ \\ \vspace{4pt} ${\scriptstyle \pm 0.0001}$} \\
\bottomrule
\end{tabular}
}
\caption{Ablation study of similarity metric and aggregation functions on Spider dataset. }
\label{table:sim_agg_study_spider}
\end{table*}

\begin{figure*}[htb]
\setlength{\tabcolsep}{-0.05cm}
\begin{tabular}{ccc}
\includegraphics[width=.35\textwidth]{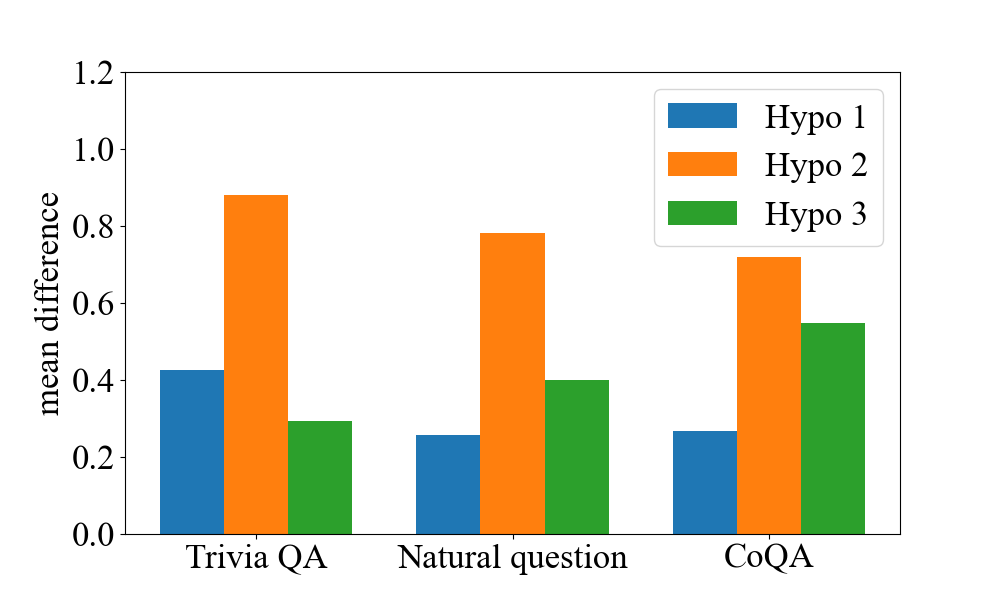}&
\includegraphics[width=.35\textwidth]{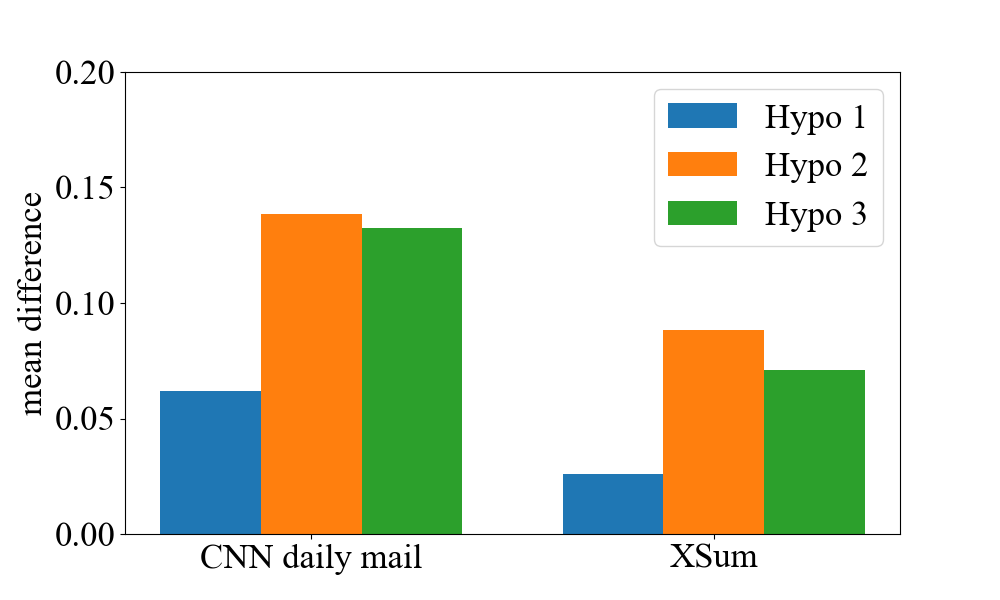}&
\includegraphics[width=.35\textwidth]{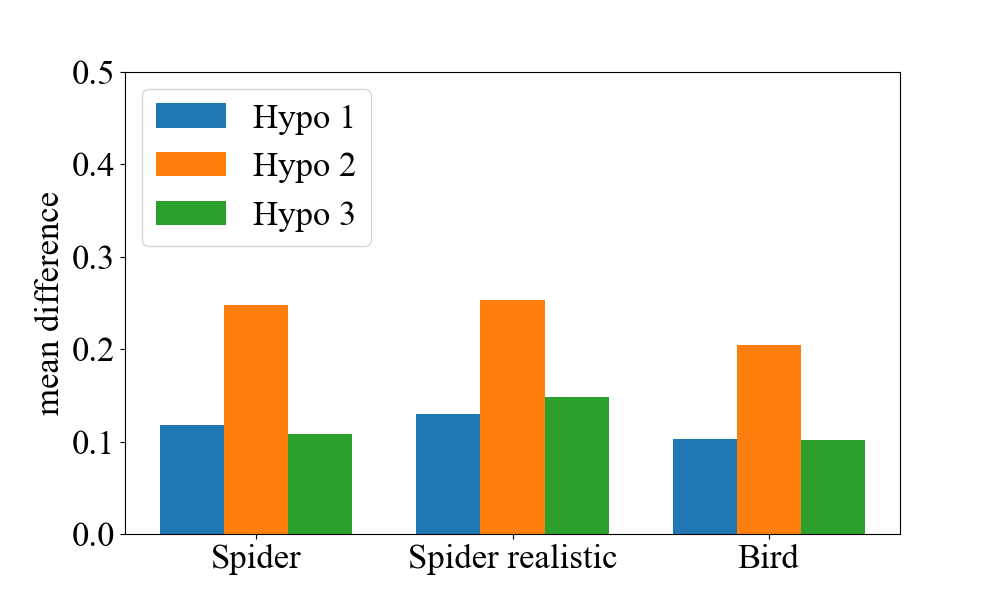}\\
{\footnotesize(a) QA} & {\footnotesize(b) Summarization }& {\footnotesize(c) Text-to-SQL}
\end{tabular}
\vspace{-0.2cm}
 \caption{Verification of hypotheses for non-default models using mean difference $\Delta \mu$  between similarity sets with Jaccard pairwise similarity on all datasets for the QA, summarization, and text-to-SQL tasks for 8 datasets. }
\label{fig:hypo_mean_diff2} 
\vspace{-0.4cm}
\end{figure*}

\begin{figure*}[htb]
\setlength{\tabcolsep}{-0.05cm}
\begin{tabular}{ccc}
\includegraphics[width=.35\textwidth]{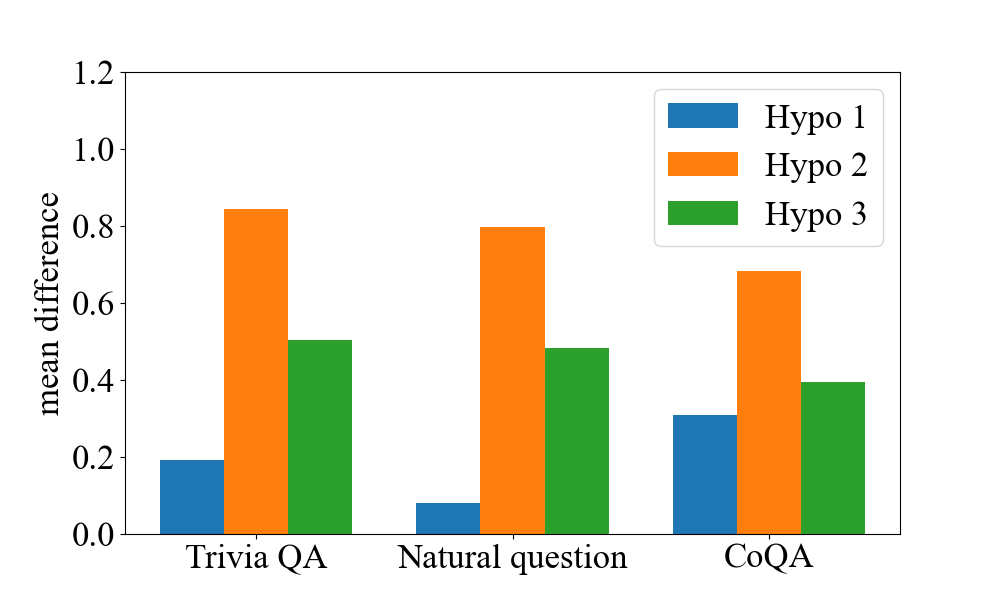}&
\includegraphics[width=.35\textwidth]{figures/mean_diff_QA.png}&
\includegraphics[width=.35\textwidth]{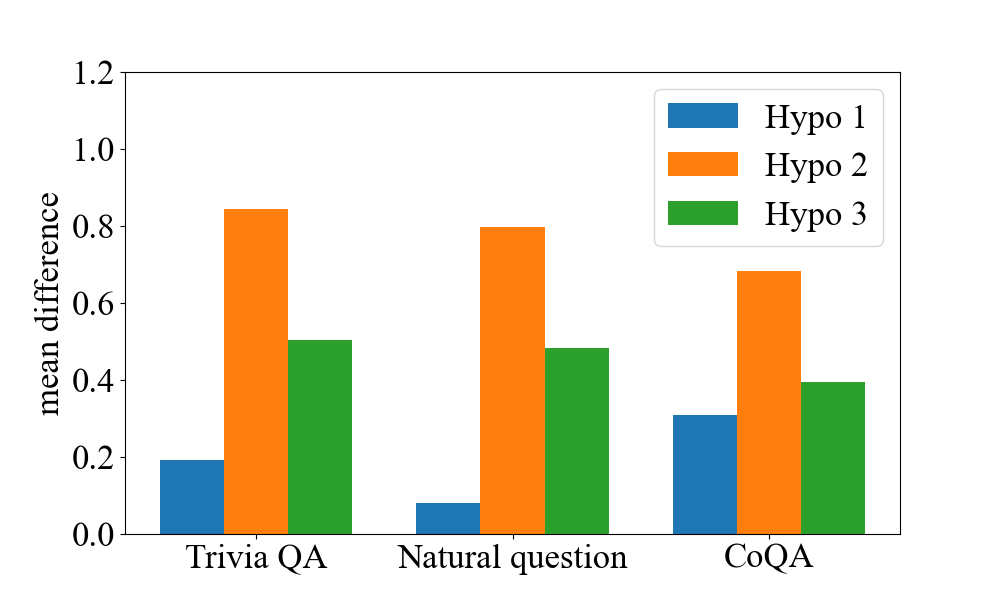}\\
{\footnotesize(a) threshold $0.4$ for QA} & {\footnotesize(b) threshold $0.5$ for QA }& {\footnotesize(c) threshold $0.6$ for QA}\\
\includegraphics[width=.35\textwidth]{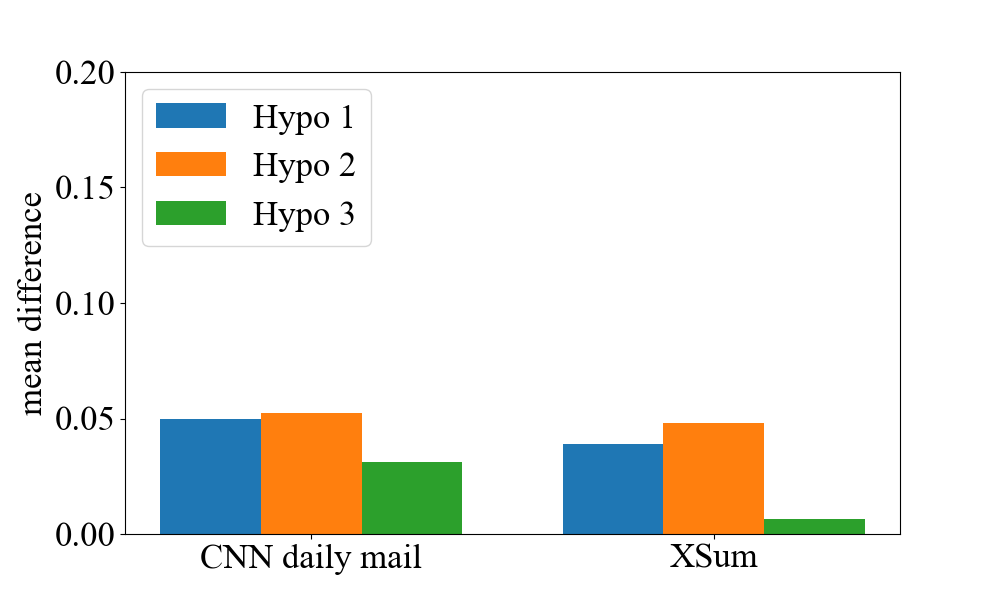}&
\includegraphics[width=.35\textwidth]{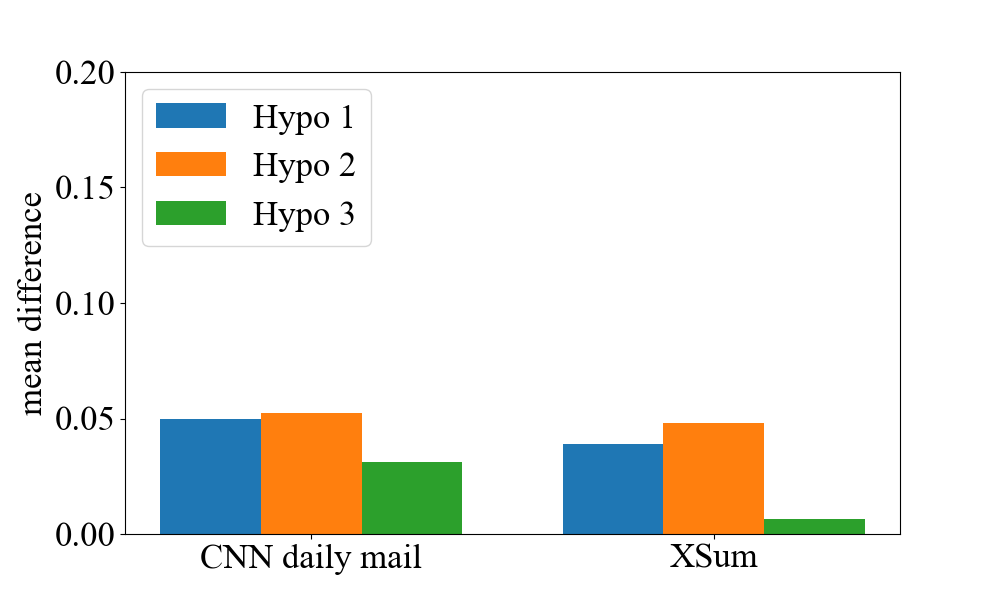}&
\includegraphics[width=.35\textwidth]{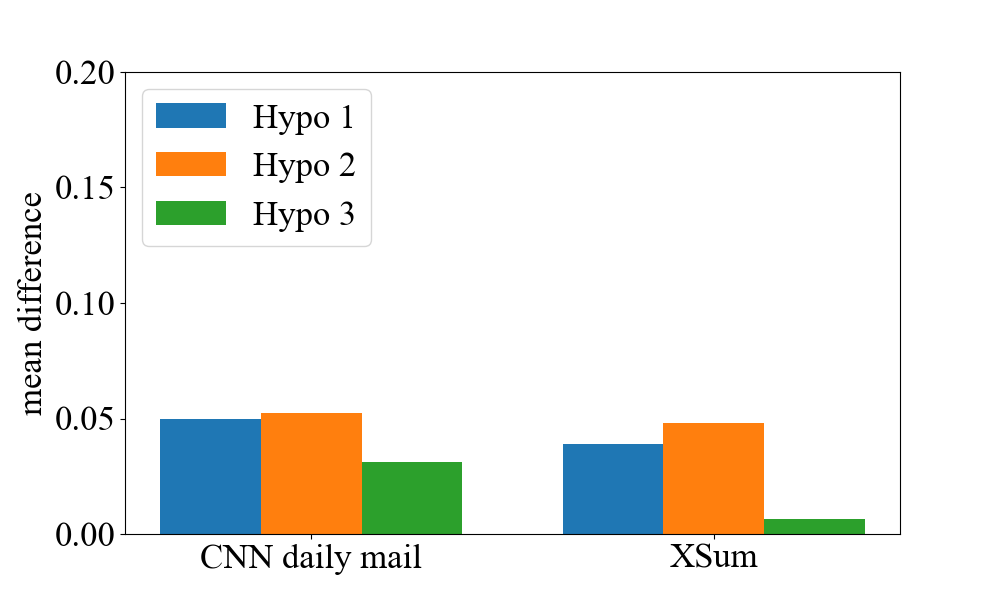}\\
{\footnotesize(a) threshold $0.1$ for summarization} & {\footnotesize(b) threshold $0.2$ for summarization}& {\footnotesize(c) threshold $0.3$ for summarization}
\end{tabular}
\vspace{-0.2cm}
 \caption{Verfication of hypotheses under varying Rouge-L score thresholds for the correctness of generations using mean difference $\Delta \mu$  between similarity sets with Jaccard arithmetic similarity on all $5$ datasets for the QA and summarization tasks. }
\label{fig:hypo_mean_diff3} 
\vspace{-0.4cm}
\end{figure*}


\subsection{Ablation Study: Similarity Metric \&
Aggregation Function}
In addition to Figure \ref{fig:agg_SQL}, 
in this section, we include the ablation study w.r.t the similarity metric and aggregation function on Spider dataset with Codellama model in Table \ref{table:sim_agg_study_spider}. In Spider dataset, some aggregation methods outperform the pairwise similarity, especially for Rouge-1 and Rouge-L score. The SQL output type similarity exhibits the least capability in capturing the viability of incorrect clusters, resulting in the least propensity for validating the consistency hypothesis. This may suggest the poor behavior of SQL output type similarity when applying to UQ methods.

\subsection{Validation of Hypotheses over Non-default Models}
To show the robustness of the validation of the proposed hypotheses across different LLM models, we plot the mean difference of correct and incorrect clusters on the dataset level for the non-default models (i.e. LLaMA 2 70 B for QA task, Mistral 8x7B for text summarization task, and fine-tuned Deepseek 33B model for text-to-SQL task ) in Figure \ref{fig:hypo_mean_diff2}. Similar to the results for the default models, all of the mean difference are positive, meaning that all of the consistency hypotheses are true in some sense. Among those, H2 is more true than H3, followed by H1. This suggests that while the validation extent varies across models, the means of the similarity sets for correct generations always exceed those of incorrect ones, indicating that the proposed consistency hypotheses hold to some degree across different models. 

\subsection{Ablation Study: Rouge-L Score}
As the correctness of generations for QA and summarization task depend on the thresholds for Rouge-L score, we conduct an ablation study over this threshold in this section. The results are shown in Figure \ref{fig:hypo_mean_diff3}, demonstrating that the extent of validation for all three versions of the hypotheses are robust to the choice of Rouge-L threshold.

\section{Visualization of the Distributions of Similarity Sets}\label{sec:dis_visualization}
\begin{figure*}[htb]
\setlength{\tabcolsep}{-0.1cm}
\begin{tabular}{ccc}
\includegraphics[width=.33\textwidth]{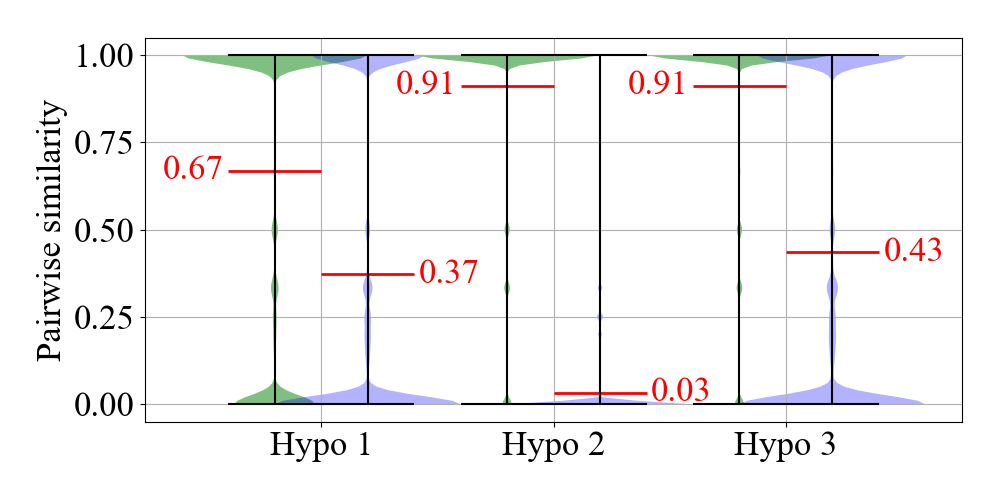}
&
\includegraphics[width=.33\textwidth]{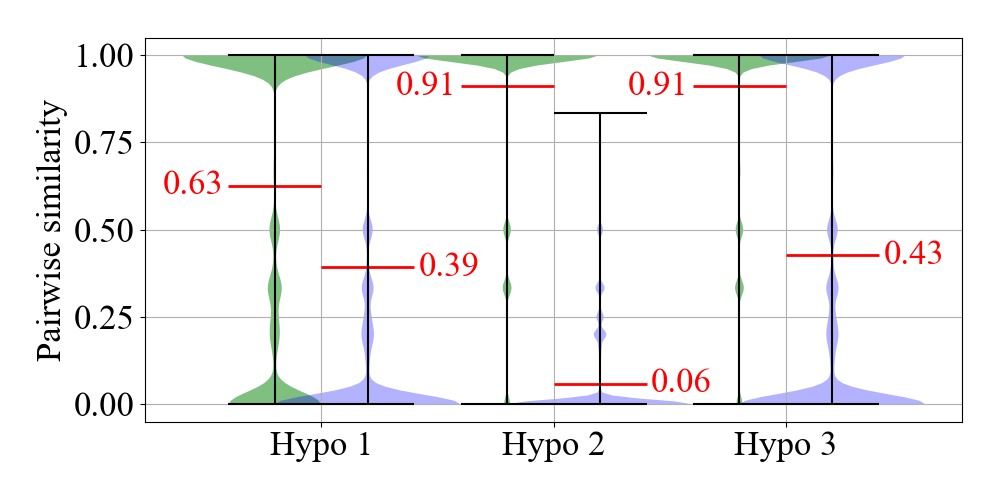}
&\includegraphics[width=.33\textwidth]{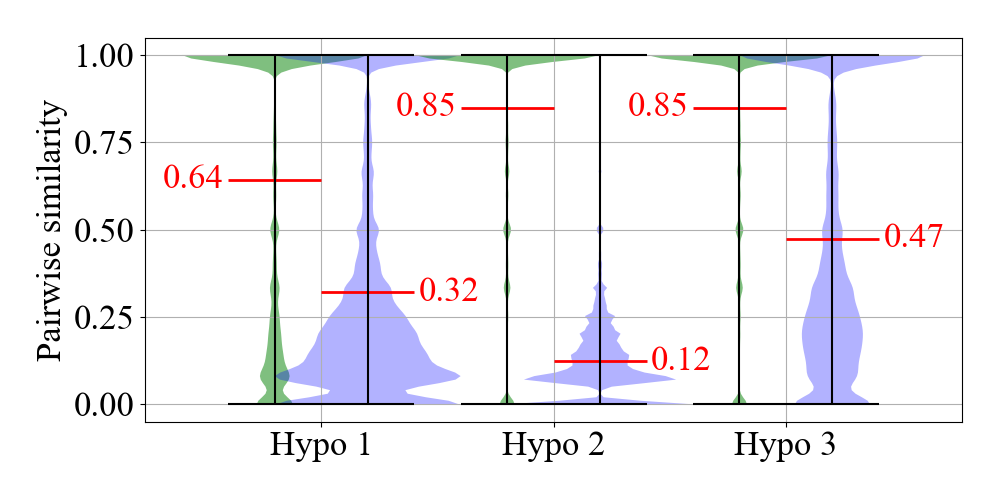}\\
{\footnotesize(a) Trivia QA dataset} & {\footnotesize(b) Natural Question dataset}& {\footnotesize(c) CoQA dataset}\\
\includegraphics[width=.33\textwidth]{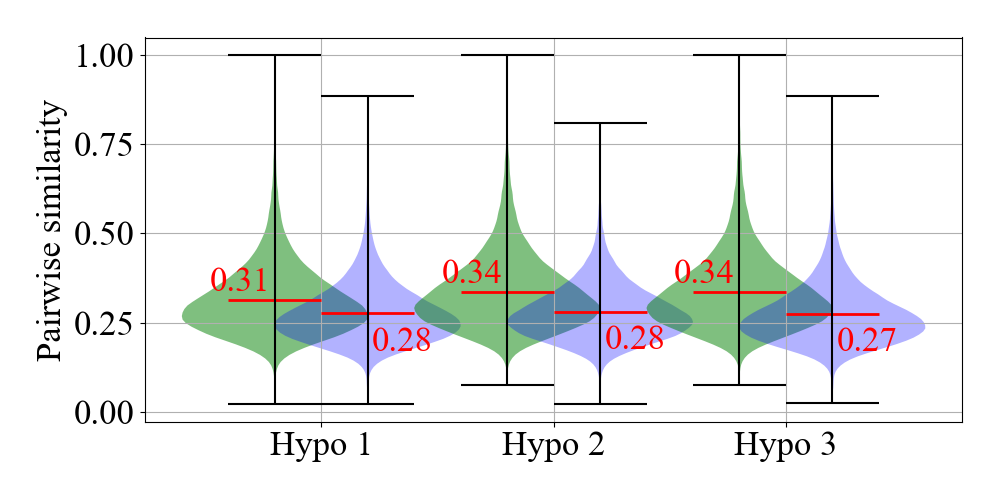}
&\includegraphics[width=.4\textwidth]{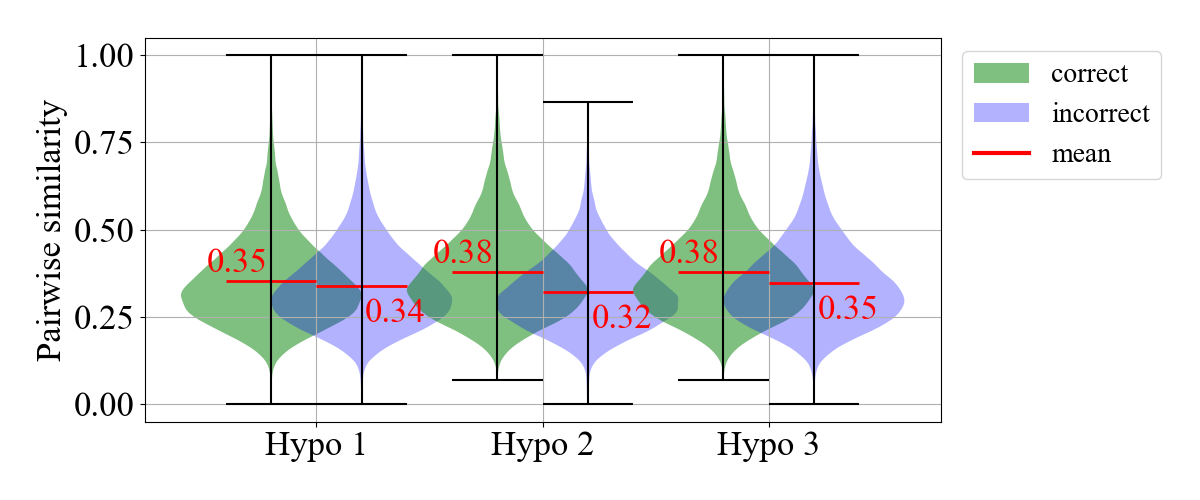}& \\
{\footnotesize(d) CNN daily mail dataset} & {\footnotesize(e) XSum dataset }& \\
\includegraphics[width=.33\textwidth]{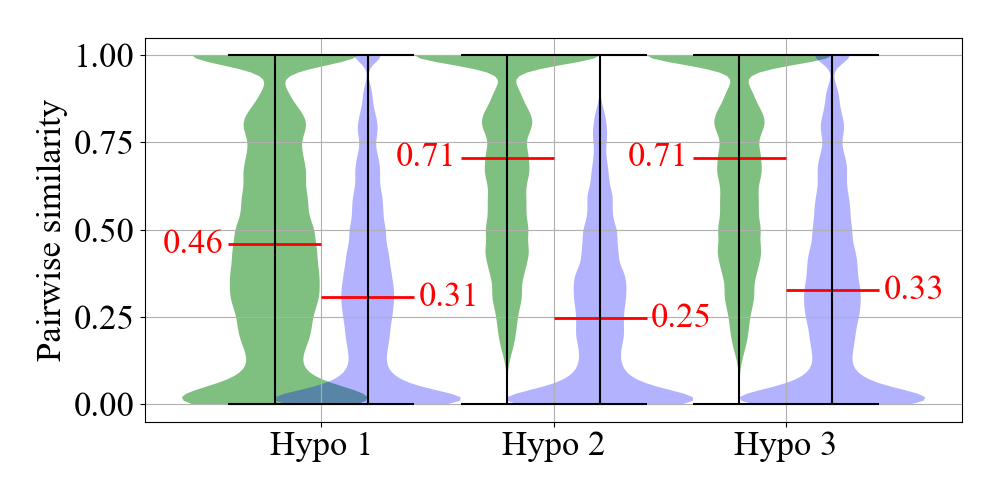}
&\includegraphics[width=.33\textwidth]{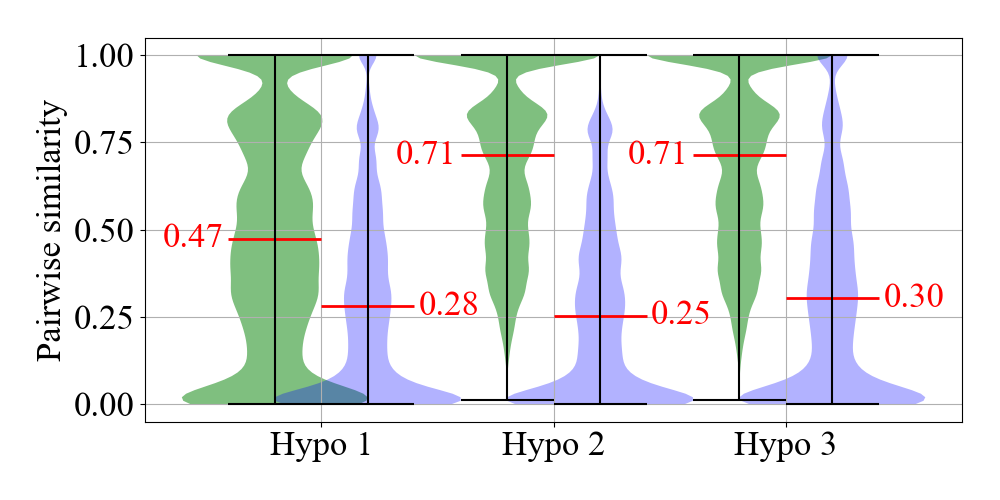}
&\includegraphics[width=.33\textwidth]{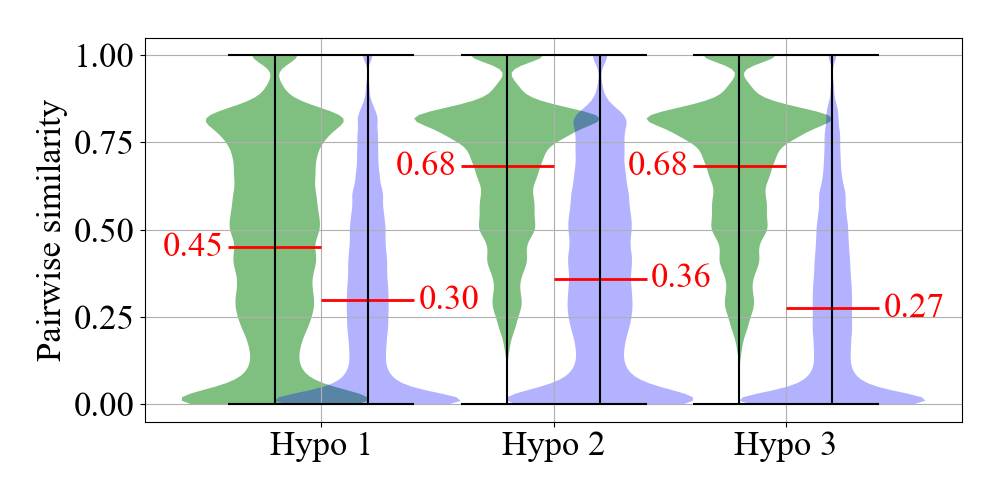}\\
{\footnotesize(f) Spider dataset} & {\footnotesize(g) Spider Realistic dataset }& {\footnotesize(h) Bird dataset}
\end{tabular}
\vspace{-0.2cm}
 \caption{Visualization of the distributions of the Jaccard pairwise similarities of correct cluster $S^C$ and incorrect cluster $S^I$ under different hypotheses on the datasets of QA, summarization, and text-to-SQL tasks. }
\label{fig:boxplot-2} 
\end{figure*}

In this section, we visualize the distributions of correct similarity sets $S^C$ and incorrect similarity sets $S^I$ across various datasets in QA, summarization, and text-to-SQL tasks in Figure \ref{fig:boxplot-2}. The mean values of pairwise similarity sets for correct generations consistently surpass those of incorrect ones across all tasks. The differentiation between these two clusters varies by task. In QA tasks, owing to the short responses, $S^C$ is concentrated around $1$, while $S^I$ centers around $0$. In text-to-SQL tasks, 
$S^C$ tends to have higher values, whereas $S^I$ is skewed towards lower values. The most challenging task from this perspective is  text summarization, where even correct answers exhibit diverse distributions. However, $S^C$ still maintains a higher mean compared to $S^I$.

This visualization of the distributions of the pairwise similarity set provides insights into the inherent difficulty levels of different tasks. It is instructive in further classifying generations as correct or incorrect through clustering based on their similarities with other generations. 

\end{document}